\documentclass[twocolumn]{article}
\usepackage{iftex}
\ifPDFTeX\pdfminorversion=7\fi
\PassOptionsToPackage{hyphens}{url}
\PassOptionsToPackage{table}{xcolor}
\usepackage{arxiv}
\AtBeginDocument{%
  \newgeometry{margin=1in,headheight=14pt,headsep=25pt,footskip=30pt}%
}
\usepackage[utf8]{inputenc}
\usepackage[T1]{fontenc}
\usepackage{latexsym}
\usepackage{microtype}
\usepackage{inconsolata}
\usepackage{xurl}
\usepackage{xcolor}
\usepackage{graphicx}
\usepackage{adjustbox}
\usepackage{caption}
\usepackage{amsmath,amssymb}
\usepackage{booktabs,seqsplit,tabularx,multirow,array}
\usepackage{algorithm,algorithmic}
\usepackage[authoryear,round]{natbib}
\renewcommand{\cite}{\citep}
\definecolor{linkblue}{RGB}{26,65,112}
\usepackage[colorlinks=true,allcolors=linkblue,breaklinks=true]{hyperref}
\renewcommand{\shorttitle}{OACR}
\hypersetup{
  pdftitle={Outcome-Anchored Credit Reallocation for Agentic Knowledge Graph Question Answering},
  pdfauthor={Mingyuan Li, Yucheng Li, Xiaxia Wang, Gong Cheng},
  pdfsubject={Agentic knowledge graph question answering},
  pdfkeywords={knowledge graph question answering, credit assignment, reinforcement learning}
}
\definecolor{PromptBg}{RGB}{235,242,252}
\definecolor{TurnBg}{RGB}{255,239,220}
\definecolor{AnswerBg}{RGB}{230,247,237}
\definecolor{GoldBlue}{RGB}{20,85,210}
\definecolor{PredRed}{RGB}{205,45,45}

\newsavebox{\arrtagbox}
\newcommand{\arrfitmath}[1]{%
  \mathord{\hbox{%
    \sbox{\arrtagbox}{\normalfont(\theequation)}%
    \adjustbox{max width=\dimexpr\columnwidth-2\wd\arrtagbox-1em\relax}{$\displaystyle #1$}%
  }}%
}
\newcommand{\arrcompactmath}[1]{%
  \arrfitmath{\thinmuskip=1mu\medmuskip=1mu\thickmuskip=1mu #1}%
}

\title{Outcome-Anchored Credit Reallocation for Agentic Knowledge Graph Question Answering}
\author{%
  Mingyuan Li\textsuperscript{1}\quad
  Yucheng Li\textsuperscript{1}\quad
  Xiaxia Wang\textsuperscript{2}\quad
  Gong Cheng\textsuperscript{1,*}\\[5pt]
  {\small \textsuperscript{1}State Key Laboratory of Novel Software Technology,}\\[-1pt]
  {\small Nanjing University, Nanjing, China}\\[2pt]
  {\small \textsuperscript{2}Department of Computer Science, University of Oxford}\\[4pt]
  {\small \textsuperscript{*}Corresponding author:
    \href{mailto:gcheng@nju.edu.cn}{\texttt{gcheng@nju.edu.cn}}}%
}

\date{}

\begin{document}
\twocolumn[{\begin{@twocolumnfalse}
\maketitle
\end{@twocolumnfalse}}]

\begin{abstract}
Agentic knowledge graph question answering (KGQA) requires large language models (LLMs) to solve complex questions through multi-turn interactions with a knowledge graph (KG). Outcome-supervised reinforcement learning derives verifiable outcome rewards from final answers but uniformly assigns each rollout's outcome-level advantage to all policy-generated tokens, creating a trajectory-to-operation mismatch since KG operations within a rollout can have varying effects on the search process and final answer. To address this, we propose Outcome-Anchored Credit Reallocation (OACR), which employs a training-only KG-state outcome model to estimate outcome-reward changes between adjacent KG search states and converts them into operation credit, from which token-wise weights are derived. These multiplicative weights reallocate token-level update magnitudes while preserving the sign of the outcome-level advantage at every generated token. Across three benchmarks, OACR achieves stronger overall performance than existing KGQA methods while using fewer inference-time LLM calls and tokens.
\end{abstract}

\section{Introduction}
% Declare the float early so it can use the first page right column.
\begin{figure}[!t]
    \centering
    \includegraphics[width=\columnwidth]{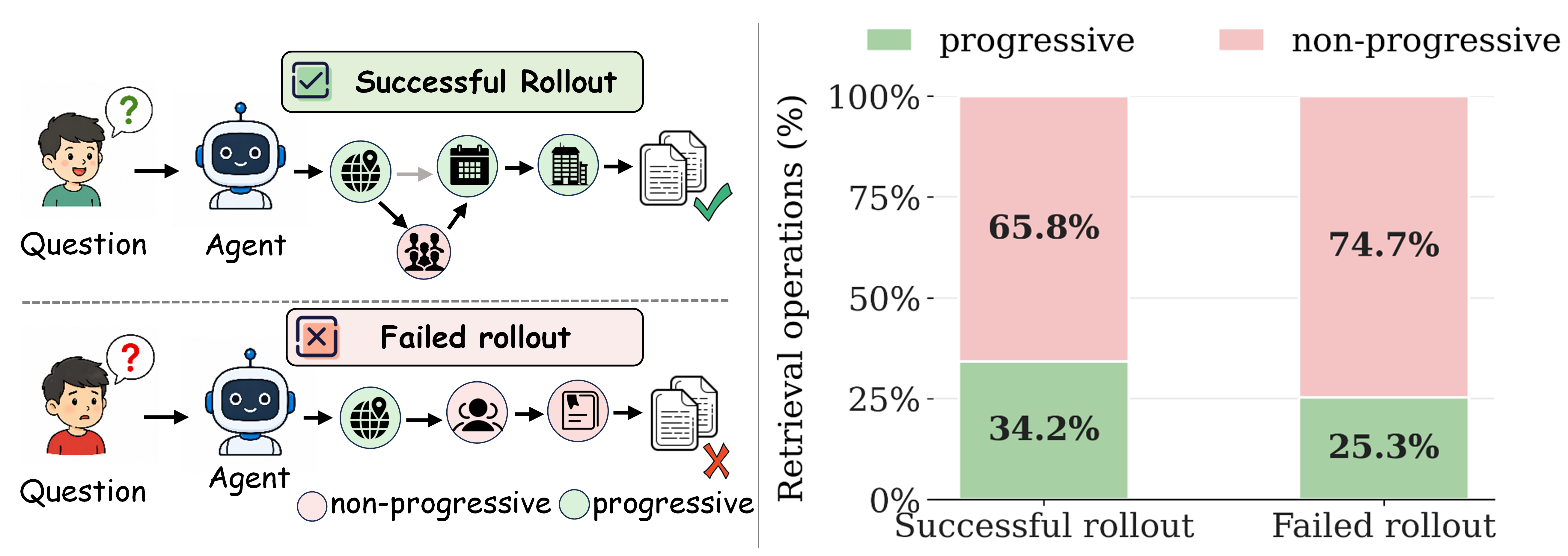}
    \caption{
    Trajectory-to-operation mismatch under outcome-supervised RL.
    Left: Successful rollouts may contain non-progressive entity-return operations, while failed rollouts may contain progressive ones.
    Right: Distribution of progressive and non-progressive entity-return operations across successful and failed rollouts, based on 5,000 such operations generated by Qwen2.5-3B on CWQ.
    }
    \label{fig:outcome_credit_issue}
\end{figure}

Knowledge graph question answering (KGQA) seeks to answer natural language questions by reasoning over facts represented as directed triples in a knowledge graph (KG)~\citep{survey1,survey2,survey3}.
Complex KGQA often requires multi-hop reasoning: a model must select relation paths that reach answer entities, follow the correct directions, and avoid irrelevant branches~\citep{rulekbqa,lightprof}.

Recent agentic methods based on large language models (LLMs) formulate KGQA as multi-turn interaction with a KG environment~\citep{tog,pog,rog,kg-r1,KnowCoder-A1,eog}.
At each turn, the agent issues a schema-valid KG operation, receives a structured observation containing returned relations or entities, and appends both to the accumulated KG search state.
This interaction process is well suited to \textit{outcome-supervised reinforcement learning} (RL): final answers provide verifiable outcome rewards, whereas labels that directly assess the utility of individual KG operations are costly to obtain and usually unavailable during training. We label an entity-return operation as \textit{progressive} if it reduces the minimum shortest-path distance between the accumulated entity set and the gold answer set, and \textit{non-progressive} otherwise. Retrieving additional gold answers or verifying constraints may leave this distance unchanged.

\begin{figure*}[t]
    \centering
    \includegraphics[width=\textwidth]{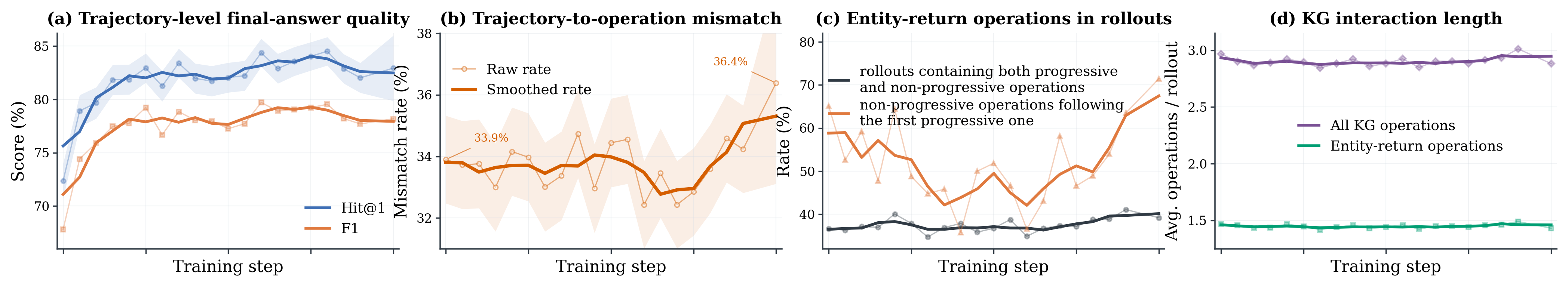}
    \caption{
    Training-process diagnostics of outcome-only GRPO in agentic KGQA.
    (a) Final-answer quality.
    (b) Raw and smoothed trajectory-to-operation mismatch rates.
    (c) Rates of rollouts containing both progressive and non-progressive entity-return operations and of non-progressive operations following the first progressive one.
    (d) Average numbers of all KG operations and entity-return operations per rollout.
    }
    \label{fig:train_process_dig}
\end{figure*}
\noindent \textbf{Motivation.}
Figure~\ref{fig:outcome_credit_issue} summarizes 5,000 entity-return
operations from rollouts generated on ComplexWebQuestions
(CWQ)~\citep{cwq} by a Qwen2.5-3B model trained with outcome-only
Group Relative Policy Optimization (GRPO)~\citep{grpo}.
In successful rollouts, 65.8\% of entity-return operations are
non-progressive, while 25.3\% in failed rollouts are progressive.
Thus, rollout outcome cannot determine whether individual
entity-return operations advance KG search.
Nevertheless, outcome-supervised RL assigns the same outcome-level
advantage to all policy-generated tokens in a rollout, including those
serializing progressive and non-progressive operations.
We call this mismatch between rollout-level supervision and individual
KG-operation effects \emph{trajectory-to-operation mismatch}.

We further examine whether outcome-only GRPO reduces this mismatch. Figure~\ref{fig:train_process_dig} reports the training-process diagnostics.
Panel (a) shows both F1 and Hit@1 rising during training, whereas Panel (b) shows the mismatch rate fluctuating between 33\% and 36\%.
Here, a mismatch is a non-progressive entity-return operation in a successful rollout or a progressive operation in a failed rollout. Panel (c) shows that the rate of rollouts containing both progressive and non-progressive entity-return operations remains at 36\% to 40\%, while the rate of non-progressive entity-return operations after the first progressive operation fluctuates between 42\% and 68\% and increases toward the end of training.
Panel (d) shows that the average interaction length remains stable at 2.9 KG operations and 1.45 entity-return operations per rollout.
\emph{Together, Figures~\ref{fig:outcome_credit_issue} and~\ref{fig:train_process_dig} show that outcome-only GRPO improves final-answer performance, yet the trajectory-to-operation mismatch persists throughout training and is not explained by changes in KG interaction length.}

These observations necessitate a credit-assignment method that \textit{distinguishes KG operations within a rollout while preserving the sign of  outcome-level advantage.}
A common alternative uses Process Reward Models (PRMs) to score intermediate steps~\citep{prm1,prm2,prm3}.
However, an operation's effect on the outcome reward depends on subsequent operations and final-answer generation, so treating these scores as independent local rewards or additive local advantages can adversely change the sign of the token-level advantage.

\noindent \textbf{Our Work.}
We propose \textit{Outcome-Anchored Credit Reallocation} (OACR) for outcome-supervised agentic KGQA.
OACR computes an outcome-level advantage from the rewards of rollouts sampled for the same question and incorporates a \textit{KG-state outcome model} to estimate outcome-reward changes between adjacent KG search states.
It converts these changes into operation credit, assigning larger weights to tokens that serialize KG operations assigned greater credit.
The token weights are normalized over all policy-generated tokens in the rollout and remain positive.
Multiplying these positive weights by the outcome-level advantage reallocates its magnitude across policy-generated tokens while preserving its sign.
The KG-state outcome model is used only during training and introduces no additional inference-time cost.
Across three benchmarks, OACR outperforms existing KGQA methods overall while using fewer inference-time LLM calls and tokens.

Our \textbf{main contributions} are as follows.
\begin{itemize}
    \item We identify and empirically analyze trajectory-to-operation mismatch under outcome-supervised RL for agentic KGQA: the final-answer outcome does not distinguish whether individual KG operations advance KG search, yet all policy-generated tokens within a rollout receive the same outcome-level advantage. This mismatch persists throughout training.

    \item We propose OACR, which converts estimated outcome-reward changes between adjacent KG search states into operation credit used to construct token weights, thereby reallocating token-level update magnitudes while preserving the sign of the outcome-level advantage.
\end{itemize}

Our code is publicly available at https://anonymous.4open.science/r/OACR/.

\section{Related Work}
Further discussion of these related methods is provided in Appendix~\ref{app:related_work}.

\noindent \textbf{LLM-based KGQA.}
LLM-based KGQA formulates reasoning as explicit KG search, enabling
LLMs to construct or revise search decisions for retrieving KG facts
~\citep{tog,rog,pog,kbqa-o1,rtsog,seal}.
\textit{These methods improve KGQA through inference-time search,
constrained decoding, or supervised trajectories, orthogonal to our OACR
that optimizes a multi-turn KG interaction policy from sampled rollouts.}

\noindent \textbf{RL for agentic KGQA.}
Recent work trains KGQA agents from sampled rollouts using outcome
rewards or reinforcement fine-tuning to optimize multi-turn KG
interaction policies~\citep{kg-r1,Graph-RFT}.
Other studies refine outcome supervision through reward modeling,
rejection sampling, or curriculum RL~\citep{eog,KnowCoder-A1}.
\textit{These methods assign rewards to complete rollouts or paths,
providing limited distinction among KG operations within a rollout and
leaving the trajectory-to-operation mismatch unresolved.
In contrast, OACR converts estimated outcome-reward changes into
operation credit for constructing token weights across KG operations.}

\noindent \textbf{Process rewards and credit assignment.}
Prior work assigns rewards or advantages to intermediate reasoning
steps or interaction turns, while preference-based optimization
reweights training examples, text segments, or tokens
~\citep{prm1,prm2,prm3,prmbut1,prmbut2,prmbut3,
reweighted1,reweighted2,reweighted3}.
In KGQA, related approaches use step-wise rewards, process reward
models, or KG schema information to guide logical-form construction
and KG reasoning~\citep{mcts-kbqa,dprm,scprm}.
\textit{These methods provide intermediate supervision, search
guidance, or optimization weights. However, accumulated intermediate rewards or
advantages can change the sign of the token-level advantage. In contrast, our OACR derives weights from outcome-reward changes between adjacent KG search states and preserves the sign of the outcome-level advantage.}

\begin{figure*}[t]
    \centering
    \includegraphics[width=\textwidth]{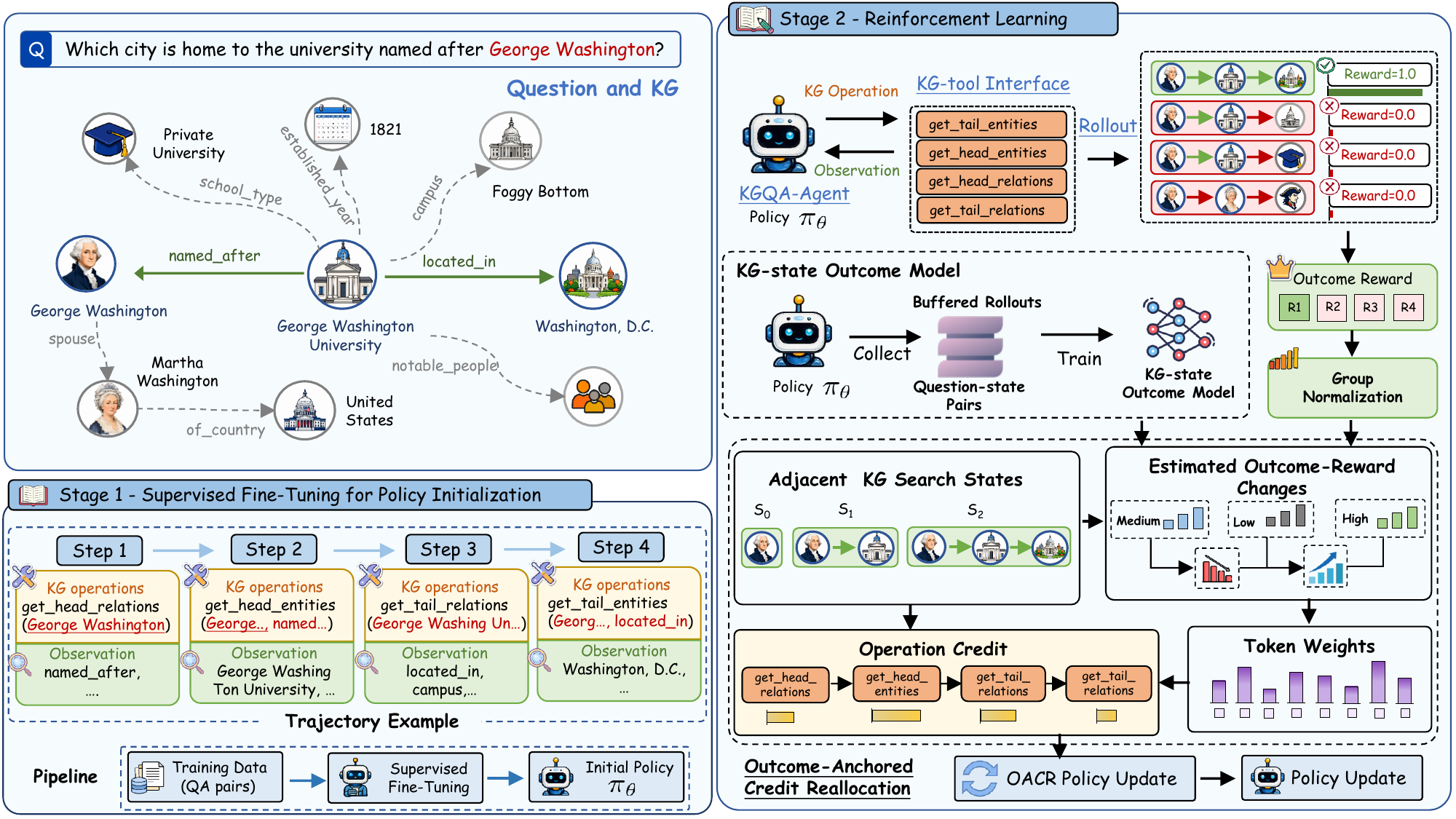}
    \caption{
    Overview of OACR.
    Supervised fine-tuning initializes a KGQA agent, followed by
outcome-supervised reinforcement learning with outcome-anchored
credit reallocation.
The KG-state outcome model is updated using recent OACR rollouts
every $U$ RL iterations and held fixed during each policy update.
The entire process is outlined in Algorithm~\ref{alg:oacr} of
Appendix~\ref{app:kg_state_outcome_model_details}.
    }
    \label{fig:framework}
\end{figure*}

\section{Method}
Figure~\ref{fig:framework} presents the two-stage OACR pipeline.
Stage 1 uses supervised fine-tuning to initialize the KGQA policy.
In Stage 2, the current policy samples rollout groups and computes an
outcome-level advantage for each rollout.
The KG-state outcome model is initially trained on outcome-only GRPO
rollouts and periodically updated using recent OACR rollouts.
Its estimates for adjacent KG search states are used to compute
operation credit and construct token weights.
These weights rescale the outcome-level advantage across
policy-generated tokens while preserving its sign.
Detailed inference rollouts are provided in
Appendix~\ref{app:oacr_inference_examples}.

\subsection{Task Formulation}
\label{sec:task_formulation}
Let $\mathcal{G}=(\mathcal{E},\mathcal{R},\mathcal{T})$ denote a KG with entity set $\mathcal{E}$, relation set $\mathcal{R}$, and directed triple set
$\mathcal{T}\subseteq\mathcal{E}\times\mathcal{R}\times\mathcal{E}$.
Each triple $(e_h,r,e_t)\in\mathcal{T}$ links head entity $e_h$ to tail entity $e_t$ through relation $r$.
Given a question $q$ with gold answer set
$\mathcal{A}_q\subseteq\mathcal{E}$, KGQA aims to identify its answer entities in $\mathcal{G}$.
Following the literature, each question is assumed to be associated with a set of mentioned start entities. In agentic KGQA, the agent interacts with the KG over multiple turns.
Let $\mathcal{O}_{\mathrm{KG}}$ denote the schema-valid KG operations available through the KG-tool interface.
At turn $t$, the agent emits an operation $a_t$, which the interface attempts to execute.
If $a_t\in\mathcal{O}_{\mathrm{KG}}$ and execution succeeds,
the returned observation lists available relations or neighboring entities.
Otherwise, the interface returns an error observation indicating a schema violation or execution error.
In both cases, the resulting observation $o_t$ is appended to the accumulated KG search state.
$S_t=(a_1,o_1,\ldots,a_t,o_t)$ with $S_0=\emptyset$.
After $L$ turns, the agent emits an ordered answer list
$\hat{\mathbf{a}}=(\hat{a}_1,\hat{a}_2,\ldots)$
conditioned on $q$ and $S_L$.
Let $\hat{\mathcal{A}}\subseteq\mathcal{E}$ denote the corresponding entity set.
The sequence
$\tau=(a_1,o_1,\ldots,a_L,o_L,\hat{\mathbf{a}})$
defines a \emph{trajectory}.
When sampled from a policy during RL or inference, the trajectory is a \emph{rollout}.

\subsection{Supervised Fine-Tuning for Policy Initialization}
\label{sec:sft}
We use supervised fine-tuning (SFT)~\citep{sft1,sft2,sft3} to initialize a KGQA policy that generates natural-language reasoning, schema-valid KG operations, and final answers conditioned on KG observations. The KG-tool interface exposes four one-hop operations.
For an entity $e\in\mathcal{E}$, the outgoing and incoming relation-list operations,
$\Gamma^{\mathrm{R}}_{\mathrm{tail}}$ and $\Gamma^{\mathrm{R}}_{\mathrm{head}}$, respectively, are
\begin{equation}
\label{eq:relation_ops}
\begin{aligned}
\Gamma^{\mathrm{R}}_{\mathrm{tail}}(e)
&=\{r\mid \exists e'\in\mathcal{E},\ (e,r,e')\in\mathcal{T}\},\\
\Gamma^{\mathrm{R}}_{\mathrm{head}}(e)
&=\{r\mid \exists e'\in\mathcal{E},\ (e',r,e)\in\mathcal{T}\}.
\end{aligned}
\end{equation}
Given a relation $r\in\mathcal{R}$, the corresponding entity-return operations are
\begin{equation}
\label{eq:entity_ops}
\begin{aligned}
\Gamma^{\mathrm{E}}_{\mathrm{tail}}(e,r)
&=\{e'\in\mathcal{E}\mid (e,r,e')\in\mathcal{T}\},\\
\Gamma^{\mathrm{E}}_{\mathrm{head}}(e,r)
&=\{e'\in\mathcal{E}\mid (e',r,e)\in\mathcal{T}\}.
\end{aligned}
\end{equation}

Each SFT trajectory follows the inference format: the model reasons in \texttt{<think>}, issues a KG operation in \texttt{<search>}, receives its observation in \texttt{<observation>}, and emits the final answer in \texttt{<answer>}.
Representative examples are provided in Appendix~\ref{app:sft_examples}.
We construct a training set
$\mathcal{D}=\{(q_i,x_i)\}_{i=1}^{N}$,
where
$x_i=(x_{i,1},\ldots,x_{i,M_i})$
is the complete serialized interaction sequence associated with the $i$-th trajectory and $M_i$ is its length.
Each trajectory is constructed by executing the KG operations corresponding to a KG path that reaches at least one gold answer entity.
The operations are executed against $\mathcal{G}$, and their observations (i.e., relation or entity lists) are inserted into the corresponding \texttt{<observation>} blocks.
Trajectories are discarded if they contain schema-invalid operations or if any entity in the \texttt{<answer>} block is absent from the previously returned entity observations (i.e., hallucinated).

Let $\pi_{\theta}$ denote the LLM policy parameterized by $\theta$.
Question and observation tokens are provided as context and therefore excluded from the training loss.
For $(q,x)\in\mathcal{D}$, let $\mathcal{I}(x)$ denote the token indices inside the \texttt{<think>}, \texttt{<search>}, and \texttt{<answer>} blocks.
The SFT training loss averages negative log-probabilities over policy-generated tokens:
\begin{equation}
\label{eq:sft_loss}
\arrfitmath{
\mathcal{L}
=
-\mathbb{E}_{(q,x)\sim\mathcal{D}}
\left[
\frac{1}{|\mathcal{I}(x)|}
\sum_{j\in\mathcal{I}(x)}
\log \pi_{\theta}(x_j\mid q,x_{<j})
\right].
}
\end{equation}

\subsection{Reinforcement Learning}
\label{sec:rl}
Although SFT teaches the policy to issue KG operations, it does not
optimize the policy using rollouts sampled from the current policy.
We therefore apply outcome-supervised RL through the KG-tool interface.
We first review GRPO under outcome supervision and then
introduce a KG-state outcome model and OACR to address its inability
to distinguish KG operations within the same rollout.

\noindent \textbf{Outcome-Supervised RL.}
Let $\mathcal{Q}_{\mathrm{train}}$ denote the empirical distribution
of questions used for RL training.
During each RL iteration, we sample
$q\sim\mathcal{Q}_{\mathrm{train}}$ and a rollout group
$\boldsymbol{\tau}=(\tau_1,\ldots,\tau_B)$ from the current policy
$\pi_{\theta_{\mathrm{old}}}$:
\begin{equation}
\label{eq:rollout_group}
\tau_i
\sim
\pi_{\theta_{\mathrm{old}}}(\cdot\mid q,\mathcal{G}),
\quad i=1,\ldots,B.
\end{equation}
Rollout $\tau_i$ has $L_i$ turns and returns answer list
$\hat{\mathbf{a}}_i$ with entity set $\hat{\mathcal{A}}_i$.
Its outcome reward combines F1 and Hit@1:
\begin{equation}
\label{eq:outcome_reward}
R_i
=
\lambda\,\mathrm{F1}(\hat{\mathcal{A}}_i,\mathcal{A}_q)
+
(1-\lambda)\,
\mathrm{Hit@1}(\hat{\mathbf{a}}_i,\mathcal{A}_q),
\end{equation}
where $\lambda\in[0,1]$ is the outcome-reward mixing coefficient
balancing answer-set coverage and top-answer accuracy.
Specifically,
\begin{equation}
\label{eq:f1_hit}
\arrcompactmath{
\mathrm{F1}(\hat{\mathcal{A}}_i,\mathcal{A}_q)
=
\frac{
2|\hat{\mathcal{A}}_i\cap\mathcal{A}_q|
}{
|\hat{\mathcal{A}}_i|+|\mathcal{A}_q|
},\quad
\mathrm{Hit@1}(\hat{\mathbf{a}}_i,\mathcal{A}_q)
=
\mathbf{1}\!\left[
\hat{a}_{i,1}\in\mathcal{A}_q
\right].
}
\end{equation}
We set $\mathrm{F1}=1$ when both $\hat{\mathcal{A}}_i$ and $\mathcal{A}_q$ are empty, and $\mathrm{Hit@1}=0$ when $\hat{\mathbf{a}}_i$ is empty. Following GRPO~\citep{grpo}, the rewards within each group are normalized to obtain an outcome-level advantage:
\begin{equation}
\label{eq:group_advantage}
A_i
=
\frac{R_i-\mu_R}{\sigma_R+\delta},
\qquad
\mu_R
=
\frac{1}{B}
\sum_{k=1}^{B}R_k,
\end{equation}
where $\sigma_R$ is the standard deviation of
$\{R_1,\ldots,R_B\}$ and $\delta>0$ is a small constant that prevents division by zero.
The sign of $A_i$ indicates whether $R_i$ is above or below the group mean.
Let $x_i$ serialize $\tau_i$, with $\mathcal{I}(x_i)$ indexing
policy-generated tokens as in Section~\ref{sec:sft}.
For each $j\in\mathcal{I}(x_i)$, the importance ratio is
\begin{equation}
\label{eq:importance_ratio}
\rho_{i,j}(\theta)
=
\frac{
\pi_{\theta}(x_{i,j}\mid q,x_{i,<j})
}{
\pi_{\theta_{\mathrm{old}}}(x_{i,j}\mid q,x_{i,<j})
}.
\end{equation}

For an importance ratio $\rho$ and an advantage $A$, we define the
clipped surrogate term as
\begin{equation}
\label{eq:clipped_surrogate}
\arrfitmath{
g(\rho,A)
=
\min\left\{
\rho A,\,
\operatorname{clip}
\left(
\rho,1-\epsilon,1+\epsilon
\right)A
\right\}.
}
\end{equation}

Here, $\epsilon\in(0,1)$ is the clipping threshold, the clipped GRPO objective is
\begin{equation}
\label{eq:grpo_loss}
\arrfitmath{
\begin{aligned}[b]
\mathcal{L}_{\mathrm{GRPO}}
&=
-\mathbb{E}_{q,\boldsymbol{\tau}}
\Bigg[
\frac{1}{B}
\sum_{i=1}^{B}
\frac{1}{|\mathcal{I}(x_i)|}
\\
&\quad
\sum_{j\in\mathcal{I}(x_i)}
g(\rho_{i,j}(\theta),A_i)
\Bigg]
+
\beta D_{\mathrm{KL}}
(\pi_{\theta}\|\pi_{\mathrm{ref}}),
\end{aligned}
}
\end{equation}

\noindent where $D_{\mathrm{KL}}(\pi_{\theta}\|\pi_{\mathrm{ref}})$ denotes the
KL divergence between the current policy $\pi_{\theta}$ and the fixed
reference policy $\pi_{\mathrm{ref}}$, and
$\beta\in[0,\infty)$ controls the strength of the KL penalty.
The expectation is over
$q\sim\mathcal{Q}_{\mathrm{train}}$ and the rollout group
$\boldsymbol{\tau}$ sampled according to Eq.~\ref{eq:rollout_group}.

\noindent \textbf{KG-state Outcome Model.} 
Eq.~\ref{eq:grpo_loss} assigns the same $A_i$ to all policy-generated
tokens in $x_i$, without distinguishing KG operations within the rollout.
We introduce a KG-state outcome model $\Phi_{\phi}$ to estimate
outcome rewards for intermediate KG search states.
We initially train $\Phi_{\phi}$ using a buffer
$\widetilde{\mathcal{T}}$ of rollouts sampled from an outcome-only
GRPO policy for questions $q\sim\mathcal{Q}_{\mathrm{train}}$.

For each buffered rollout
$\tilde{\tau}_m\in\widetilde{\mathcal{T}}$, let $q_m$ denote its
question, $\tilde{R}_m$ its outcome reward from
Eq.~\ref{eq:outcome_reward}, and $\tilde{S}_{m,t}$ its accumulated
KG search state after $t$ turns, with $\tilde{S}_{m,0}=\emptyset$.
For each canonicalized question-state pair $(q,S)$ observed in the
current buffer $\widetilde{\mathcal{T}}$, we define $y(q,S)$ as
the mean outcome reward of the retained rollouts that visit that pair.
Canonicalization preserves operation order and normalizes serialized
fields and returned observations, as detailed in
Appendix~\ref{app:state-repeat}.
The fitting loss is
\begin{equation}
\label{eq:kg_state_fit_loss}
\mathcal{L}_{\mathrm{fit}}
=
\mathbb{E}_{(q,S)}
\left[
\left(
\Phi_{\phi}(q,S)-y(q,S)
\right)^2
\right].
\end{equation}

Because operation credit depends on differences between adjacent
outcome-reward estimates, their ordering directly affects token
weights.
We therefore add a temporal ordering regularizer weighted by the
rollout outcome reward.
For two states from the same buffered rollout with $s<t$, we define
\begin{equation}
\label{eq:kg_state_order_loss}
\arrfitmath{
\begin{aligned}[b]
\mathcal{L}_{\mathrm{ord}}
&=
\mathbb{E}_{m,\,s<t}
\Big[
-\tilde{R}_m
\log \sigma\!\Big(
\\
&\quad\Phi_{\phi}(q_m,\tilde{S}_{m,t})
-
\Phi_{\phi}(q_m,\tilde{S}_{m,s})
\Big)
\Big],
\end{aligned}
}
\end{equation}
where $\sigma(\cdot)$ is the sigmoid function,
$L_m$ is the number of turns in buffered rollout $\tilde{\tau}_m$,
and the expectation averages over all state pairs satisfying
$0\leq s<t\leq L_m$ within that rollout.
This regularizer softly favors higher estimates for later states,
with larger weights for rollouts receiving higher outcome rewards.
The final training objective is
\begin{equation}
\label{eq:kg_state_outcome_loss}
\mathcal{L}_{\Phi}
=
\mathcal{L}_{\mathrm{fit}}
+
\eta\,\mathcal{L}_{\mathrm{ord}},
\end{equation}
where $\eta\geq0$ is the temporal ordering regularization strength. 

Policy updates can change both the KG search states visited
and the expected outcome reward after visiting a given state.
Before each policy update, we compute and cache the outcome-reward
estimates and token weights using the current $\Phi_{\phi}$.
The model and cached quantities remain fixed throughout the update,
and the policy loss does not propagate gradients through them.
After the policy update, we add the sampled rollouts and their
observed outcome rewards to $\widetilde{\mathcal{T}}$, discarding
the oldest rollouts when its capacity is exceeded.
Every $U$ RL iterations, we recompute $y(q,S)$ from the current
buffer and continue training $\Phi_{\phi}$ using
Eq.~\ref{eq:kg_state_outcome_loss}.
The updated model is used for subsequent newly sampled rollouts.
The KG-state outcome model is used only during training.
Algorithms~\ref{alg:kg_state_outcome_model} and~\ref{alg:oacr}
in Appendix~\ref{app:kg_state_outcome_model_details} summarize
model fitting and OACR training, respectively.

\noindent \textbf{Outcome-Anchored Credit Reallocation.}
Although $\Phi_{\phi}(q,S_{i,t})$ estimates the outcome reward of
the accumulated KG search state after operation $a_{i,t}$, directly
using this absolute KG-state outcome estimate would also reflect all
preceding KG operations contained in $S_{i,t}$.
OACR therefore uses the difference between the estimated outcome
rewards of adjacent KG search states to distinguish the change after
$a_{i,t}$.
Specifically, we define the clipped estimated outcome-reward change as
\begin{equation}
\label{eq:estimated_outcome_reward_change}
\arrfitmath{
d_{i,t}
=
\operatorname{clip}
\left(
\Phi_{\phi}(q,S_{i,t})
-
\Phi_{\phi}(q,S_{i,t-1}),
-c,c
\right),
}
\end{equation}
where $c>0$ is the estimated outcome-reward change bound.
A positive $d_{i,t}$ indicates an increase in the estimated outcome
reward, while a negative value indicates a decrease.
These changes are used only to construct operation credit.

OACR conditions operation credit on the sign of $A_i$.
For $A_i>0$, credit is assigned according to the positive part of $d_{i,t}$.
For $A_i<0$, it is determined by the negative part of $d_{i,t}$ and whether the operation is invalid:
\begin{equation}
\label{eq:outcome_conditioned_operation_credit}
b_{i,t}
=
\begin{cases}
\max(d_{i,t},0),
& A_i>0,\\
\max(-d_{i,t},0)
+
\kappa\,\mathbf{1}^{\mathrm{inv}}_{i,t},
& A_i<0,\\
0,
& A_i=0,
\end{cases}
\end{equation}
where $\mathbf{1}^{\mathrm{inv}}_{i,t}$ indicates whether
$a_{i,t}$ is a schema-invalid KG operation or fails to execute,
and $\kappa\geq0$ controls the additional operation-credit magnitude
assigned to such operations.
Thus, $b_{i,t}\geq0$, while $A_i$ alone determines the sign of the
token-level advantage. Let $\mathcal{J}_{i,t}\subseteq\mathcal{I}(x_i)$ denote the indices of the policy-generated tokens that serialize $a_{i,t}$ in the corresponding \texttt{<search>} block.
For each $j\in\mathcal{I}(x_i)$, the reallocated token weight is
\begin{equation}
\label{eq:raw_token_weight}
\arrfitmath{
\omega_{i,j}
=
\begin{cases}
1+\alpha b_{i,t},
& j\in\mathcal{J}_{i,t}\ \text{for some }t,\\
1,
& j\in\mathcal{I}(x_i)\setminus
\bigcup_{t=1}^{L_i}\mathcal{J}_{i,t}.
\end{cases}
}
\end{equation}
where $\alpha\geq0$ is the reallocation strength.
We normalize these weights over all policy-generated tokens:
\begin{equation}
\label{eq:normalized_token_weight}
w_{i,j}
=
\frac{
\omega_{i,j}
}{
\frac{1}{|\mathcal{I}(x_i)|}
\sum_{\ell\in\mathcal{I}(x_i)}
\omega_{i,\ell}
}.
\end{equation}
The token-level advantage $\widetilde{A}_{i,j}=A_iw_{i,j}$
retains the sign of $A_i$ because $w_{i,j}>0$.
Substituting it for $A_i$ in Eq.~\ref{eq:grpo_loss} gives:
\begin{equation}
\label{eq:oacr_loss}
\arrfitmath{
\begin{aligned}[b]
\mathcal{L}_{\mathrm{OACR}}
&=
-\mathbb{E}_{q,\boldsymbol{\tau}}
\Bigg[
\frac{1}{B}
\sum_{i=1}^{B}
\frac{1}{|\mathcal{I}(x_i)|}
\\
&\quad
\sum_{j\in\mathcal{I}(x_i)}
g(\rho_{i,j}(\theta),\widetilde{A}_{i,j})
\Bigg]
+
\beta D_{\mathrm{KL}}
(\pi_{\theta}\|\pi_{\mathrm{ref}}).
\end{aligned}
}
\end{equation}
Operation credit reweights outcome supervision; the KL penalty
retains the GRPO coefficient $\beta$ and is not reweighted.

\section{Experiments}
\paragraph{Datasets and Metrics.}
We evaluate OACR with Qwen2.5-3B, Qwen2.5-7B, and Llama-3.1-8B
on WebQSP~\citep{webqsp}, CWQ~\citep{cwq}, and
GrailQA~\citep{grailqa}.
Following prior work~\citep{KnowCoder-A1,dprm,eog}, we report
F1 and Hit@1.
Inference costs are measured by mean LLM calls, input tokens,
output tokens, and latency per question.
Dataset statistics are provided in Appendix~\ref{dataset_stats},
and experimental settings and hyperparameters in
Appendices~\ref{experimental_settings}
and~\ref{app:kg_state_outcome_model_details}.

\label{sec:experiments}
\begin{table}[t]
\centering
{%
\small
\setlength{\tabcolsep}{1mm}
\begin{adjustbox}{max width=\columnwidth}
\begin{tabular}{@{}lcccccc@{}}
\toprule
\multirow{2}{*}{\textbf{Method}}
& \multicolumn{2}{c}{\textbf{CWQ}}
& \multicolumn{2}{c}{\textbf{WebQSP}}
& \multicolumn{2}{c}{\textbf{GrailQA}} \\
\cmidrule(lr){2-3}
\cmidrule(lr){4-5}
\cmidrule(lr){6-7}
& \textbf{F1} & \textbf{Hit@1}
& \textbf{F1} & \textbf{Hit@1}
& \textbf{F1} & \textbf{Hit@1} \\
\midrule

\rowcolor[HTML]{F2F2F2}
\multicolumn{7}{c}{\textit{Training-free Methods}} \\

Qwen2.5-3B
& 17.7 & 22.0 & 30.3 & 44.6 & 11.6 & 13.8 \\

Qwen2.5-7B
& 20.7 & 25.7 & 33.2 & 50.9 & 13.6 & 15.9 \\

Llama-3.1-8B
& 22.4 & 28.1 & 34.8 & 55.5 & 17.7 & 21.3 \\

Gemini-2.5-Pro
& -- & 65.3 & -- & 79.8 & -- & -- \\

GPT-5
& -- & 67.6 & -- & 77.5 & -- & -- \\

ToG
& -- & 68.5 & -- & 82.6 & -- & 81.4 \\

PoG
& -- & 75.0 & -- & 87.3 & -- & 84.7 \\

ReknoS
& -- & 66.8 & -- & 83.8 & -- & 80.5 \\

MCTS-KBQA
& 64.87 & -- & 72.49 & -- & -- & -- \\

\midrule
\rowcolor[HTML]{F2F2F2}
\multicolumn{7}{c}{\textit{Training-based Methods}} \\

\multicolumn{7}{c}{
\textcolor[HTML]{6B7280}{
\textit{\textemdash\ backbone: Qwen2.5-3B \textemdash}
}
} \\

KG-R1
& -- & 65.3 & -- & 82.8 & -- & -- \\

\rowcolor[HTML]{F4F8FD}
OACR$_{\mathrm{SFT}}$
& 64.9
& 70.1
& 79.6
& 86.8
& 72.6
& 79.0 \\

\rowcolor[HTML]{F4F8FD}
OACR$_{\mathrm{GRPO}}$
& \underline{68.1}
& \underline{74.1}
& \underline{80.6}
& \underline{87.8}
& \underline{73.8}
& \underline{81.3} \\

\rowcolor[HTML]{DDEBFF}
\textbf{OACR}
& \textbf{71.0}
& \textbf{76.4}
& \textbf{81.5}
& \textbf{88.5}
& \textbf{75.4}
& \textbf{82.3} \\

\multicolumn{7}{c}{
\textcolor[HTML]{6B7280}{
\textit{\textemdash\ backbone: Qwen2.5-7B \textemdash}
}
} \\

RoG
& -- & 46.3 & -- & 79.9 & -- & -- \\

Graph-RFT
& -- & 78.4 & -- & 87.3 & -- & -- \\

KnowCoder-A1
& 68.3 & 75.7 & 77.2 & 80.1 & -- & 80.5 \\

EoG
& 73.8
& 82.7
& 78.1
& \underline{90.7}
& \textbf{88.5}
& \textbf{91.7} \\

\rowcolor[HTML]{F4F8FD}
OACR$_{\mathrm{SFT}}$
& 68.6
& 73.0
& 82.4
& 89.0
& 75.3
& 81.3 \\

\rowcolor[HTML]{F4F8FD}
OACR$_{\mathrm{GRPO}}$
& \underline{75.5}
& \underline{82.9}
& \underline{85.7}
& 90.5
& 80.6
& 86.2 \\

\rowcolor[HTML]{DDEBFF}
\textbf{OACR}
& \textbf{80.1}
& \textbf{83.5}
& \textbf{86.2}
& \textbf{91.3}
& \underline{82.5}
& \underline{88.1} \\

\multicolumn{7}{c}{
\textcolor[HTML]{6B7280}{
\textit{\textemdash\ backbone: Llama-3.1-8B \textemdash}
}
} \\

KBQA-o1
& -- & -- & 57.8 & -- & 78.5 & -- \\

EoG
& \underline{77.9}
& \textbf{86.6}
& 81.3
& \textbf{92.8}
& \textbf{90.6}
& \textbf{92.1} \\

DPRM
& 59.9
& 73.0
& 72.3
& 87.2
& --
& -- \\

SCPRM
& --
& 73.3
& --
& --
& --
& -- \\

\rowcolor[HTML]{F4F8FD}
OACR$_{\mathrm{SFT}}$
& 70.0
& 74.7
& 82.5
& 89.1
& 76.0
& 82.7 \\

\rowcolor[HTML]{F4F8FD}
OACR$_{\mathrm{GRPO}}$
& 76.8
& 84.6
& \underline{86.2}
& 90.7
& 81.3
& 87.2 \\

\rowcolor[HTML]{DDEBFF}
\textbf{OACR}
& \textbf{81.4}
& \underline{85.2}
& \textbf{86.7}
& \underline{91.5}
& \underline{83.2}
& \underline{89.1} \\

\bottomrule
\end{tabular}
\end{adjustbox}
}

\caption{
Overall performance on CWQ, WebQSP, and GrailQA.
Boldface and underlining denote the best and second-best results
within each backbone-matched comparison block,
-- denotes an unreported result. 
}
\label{tab:main_results}
\end{table}

\begin{table}[t]
\centering
{%
\small
\setlength{\tabcolsep}{1mm}
\begin{adjustbox}{max width=\columnwidth}
\begin{tabular}{@{}lcccccc@{}}
\toprule
\multirow{2}{*}{\textbf{Variant}}
& \multicolumn{2}{c}{\textbf{CWQ}}
& \multicolumn{2}{c}{\textbf{WebQSP}}
& \multicolumn{2}{c}{\textbf{GrailQA}} \\
\cmidrule(lr){2-3}
\cmidrule(lr){4-5}
\cmidrule(lr){6-7}
& \textbf{F1} & \textbf{Hit@1}
& \textbf{F1} & \textbf{Hit@1}
& \textbf{F1} & \textbf{Hit@1} \\
\midrule

\rowcolor[HTML]{F2F2F2}
\multicolumn{7}{l}{\textit{Training settings}} \\
OACR$_{\mathrm{SFT}}$
& 68.6 & 73.0 & 82.4 & 89.0 & 75.3 & 81.3 \\
Outcome-only GRPO
& 75.5 & 82.9 & 85.7 & 90.5 & 80.6 & 86.2 \\
\rowcolor[HTML]{DDEBFF}
\textbf{OACR}
& \textbf{80.1} & \textbf{83.5}
& \textbf{86.2} & \textbf{91.3}
& \textbf{82.5} & \textbf{88.1} \\

\midrule
\rowcolor[HTML]{F2F2F2}
\multicolumn{7}{l}{\textit{OACR variants}} \\
\textit{Absolute state est.}
& 77.2 & 82.8 & 85.9 & 90.6 & 81.3 & 86.6 \\
\textit{Additive local adv.}
& 74.8 & 81.5 & 85.1 & 89.9 & 80.0 & 85.4 \\
w/o \textit{ordering reg.}
& 79.0 & 83.1 & 86.0 & 91.0 & 81.9 & 87.4 \\
\bottomrule
\end{tabular}
\end{adjustbox}
}

\caption{
Ablation study of OACR on Qwen2.5-7B.
\emph{Absolute state est.}, \emph{Additive local adv.}, and
\emph{ordering reg.} abbreviate absolute KG-state outcome estimate,
additive local advantages, and temporal ordering regularizer.
}
\label{tab:ablation}
\end{table}

\paragraph{Baselines.}
\emph{Training-free baselines} include direct prompting with
Qwen2.5-3B, Qwen2.5-7B, Llama-3.1-8B, Gemini-2.5-Pro, and GPT-5,
as well as inference-time KG search and planning methods
ToG~\citep{tog}, PoG~\citep{pog}, ReknoS~\citep{ReknoS},
and MCTS-KBQA~\citep{mcts-kbqa}.
\emph{Training-based baselines} use SFT or RL for KGQA and include
KG-R1~\citep{kg-r1}, RoG~\citep{rog}, KBQA-o1~\citep{kbqa-o1},
Graph-RFT~\citep{Graph-RFT}, KnowCoder-A1~\citep{KnowCoder-A1},
EoG~\citep{eog}, DPRM~\citep{dprm}, and SCPRM~\citep{scprm}.
OACR$_{\mathrm{SFT}}$ denotes the SFT-initialized policy.
OACR$_{\mathrm{GRPO}}$ further optimizes this policy using
outcome-only GRPO, while OACR additionally applies
outcome-anchored credit reallocation.

\paragraph{Main Results.}
OACR improves both F1 and Hit@1 over OACR$_{\mathrm{SFT}}$
across all three datasets and backbones
(Table~\ref{tab:main_results}; split-wise GrailQA results are
provided in Appendix~\ref{app:grailqa_split_results}).
It also outperforms OACR$_{\mathrm{GRPO}}$ in both metrics
on all three datasets with Qwen2.5-3B and Qwen2.5-7B.
The largest F1 gains over OACR$_{\mathrm{GRPO}}$ occur on CWQ,
reaching 2.9 and 4.6 points for the two backbones, respectively.
Among reported methods using the same backbone, OACR with
Qwen2.5-3B outperforms KG-R1 in Hit@1 on CWQ and WebQSP.
On these two datasets, OACR with Qwen2.5-7B ranks first in
both metrics, while OACR with Llama-3.1-8B ranks first in F1
and second in Hit@1.
With these two backbones, OACR ranks second to EoG in both
metrics on GrailQA.
\textit{The gains over OACR$_{\mathrm{GRPO}}$ support OACR's
use of estimated outcome-reward changes to construct
multiplicative token weights that reallocate the outcome-level
advantage across KG operations, improving final-answer quality
beyond outcome-only GRPO.}

\paragraph{Ablation Study.}
Using absolute KG-state outcome estimates
(\emph{Absolute state est.}) or removing the temporal ordering
regularizer while retaining outcome-reward fitting
(\emph{w/o ordering reg.}) lowers F1 and Hit@1 across all
three datasets (Table~\ref{tab:ablation}).
Absolute estimates reflect the accumulated KG search state,
whereas differences between adjacent states estimate the change
following each KG operation.
Even with these changes retained, replacing multiplicative
token weights with additive local advantages
(\emph{Additive local adv.}) causes the largest decrease among
the variants, lowering CWQ F1 by 5.3 points.
\textit{These results show that OACR improves final-answer
quality by distinguishing KG operations while preserving
the sign of the outcome-level advantage.}

\begin{figure}[t]
    \centering
    \includegraphics[width=\columnwidth]{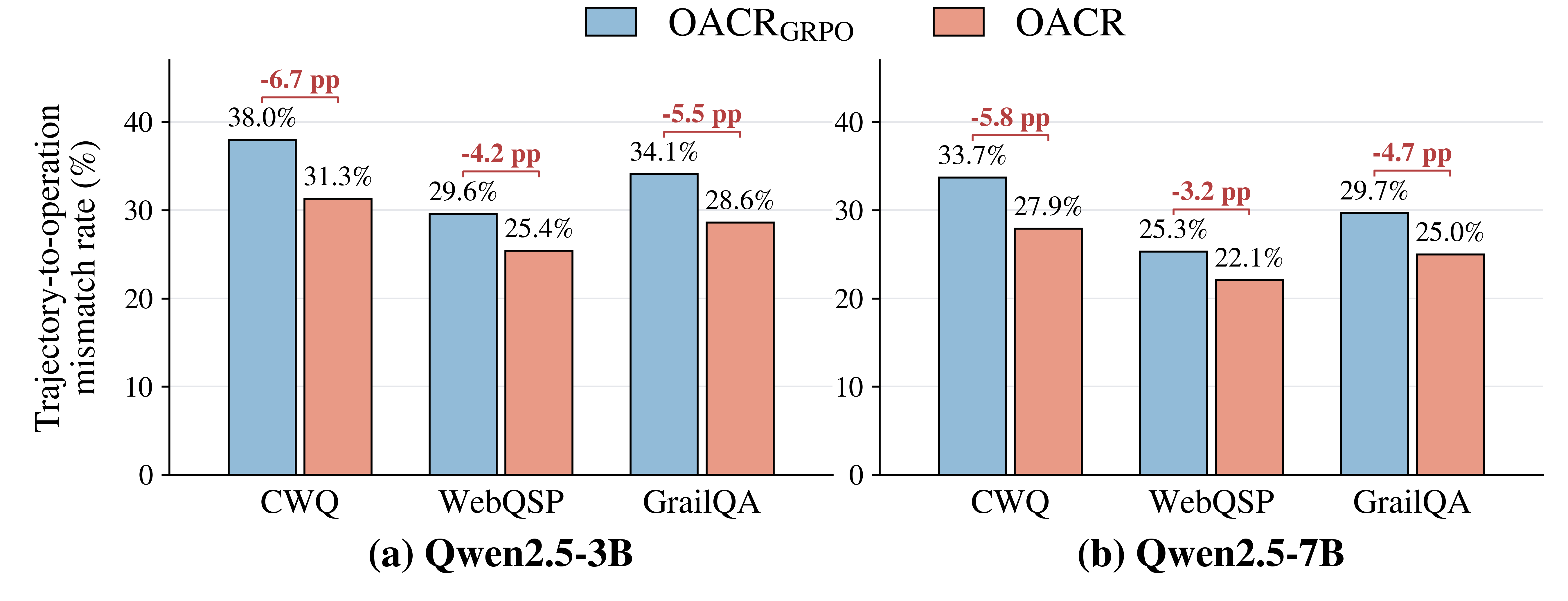}
    \caption{Trajectory-to-operation mismatch rates of
OACR$_{\mathrm{GRPO}}$ and OACR on CWQ, WebQSP, and GrailQA.
\textbf{Left}: Qwen2.5-3B; \textbf{Right}: Qwen2.5-7B.}
\label{mismatch_comparison}
\end{figure}

\paragraph{Mismatch Analysis.}
Figure~\ref{mismatch_comparison} compares the
trajectory-to-operation mismatch rates of OACR and
OACR$_{\mathrm{GRPO}}$ across three datasets using
Qwen2.5-3B and Qwen2.5-7B.
OACR consistently achieves lower mismatch rates across
all settings, with the largest reductions observed on CWQ:
6.7 and 5.8 percentage points for Qwen2.5-3B and
Qwen2.5-7B, respectively.
\textit{These results show that OACR mitigates the
trajectory-to-operation mismatch that persists under
outcome-only GRPO while improving final-answer quality.}

\paragraph{Retrieval Coverage Analysis.}
We evaluate gold-answer retrieval before final-answer generation
using Partial Coverage (PC) and Exact Coverage (EC), with
definitions and full results in
Appendix~\ref{app:retrieval_coverage_analysis}.
With Qwen2.5-3B, OACR achieves the highest PC and EC among
the compared methods across all three datasets.
On CWQ, PC and EC reach 90.1\% and 87.9\%.
Their small gap indicates that most questions with at least
one retrieved gold answer also have the complete gold answer
set in their KG observations.
Unlike F1 and Hit@1, these measures do not depend on
the generated answer list.
\textit{These results show that OACR retrieves complete gold
answer sets for more questions, providing the entities needed
for final-answer generation.}

\begin{figure}[t]
    \centering
    \includegraphics[width=\columnwidth]{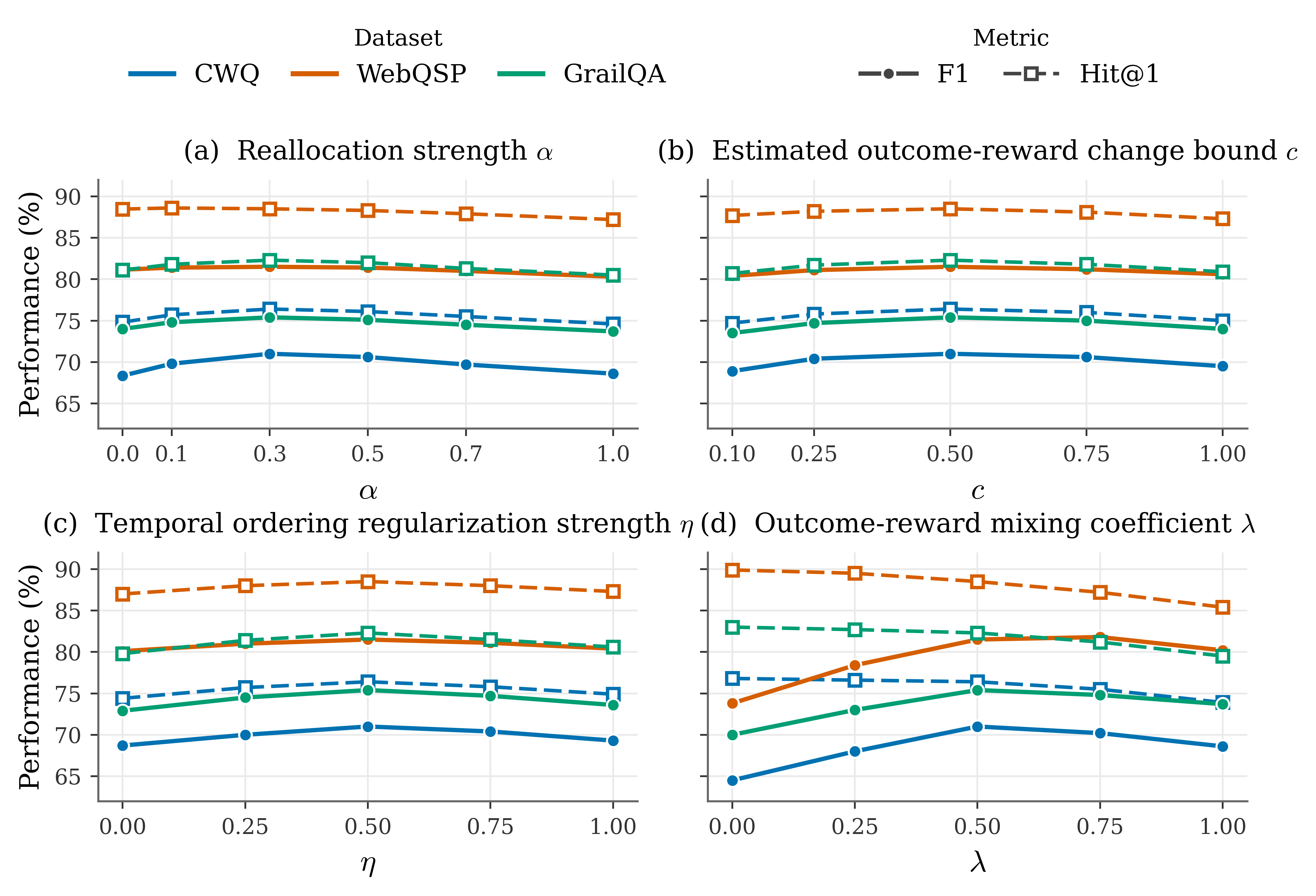}
    \caption{Hyperparameter analysis on Qwen2.5-3B.}
    \label{fig:hyperparameter_sensitivity}
\end{figure}

\begin{figure}[t]
    \centering
    \includegraphics[width=\columnwidth]{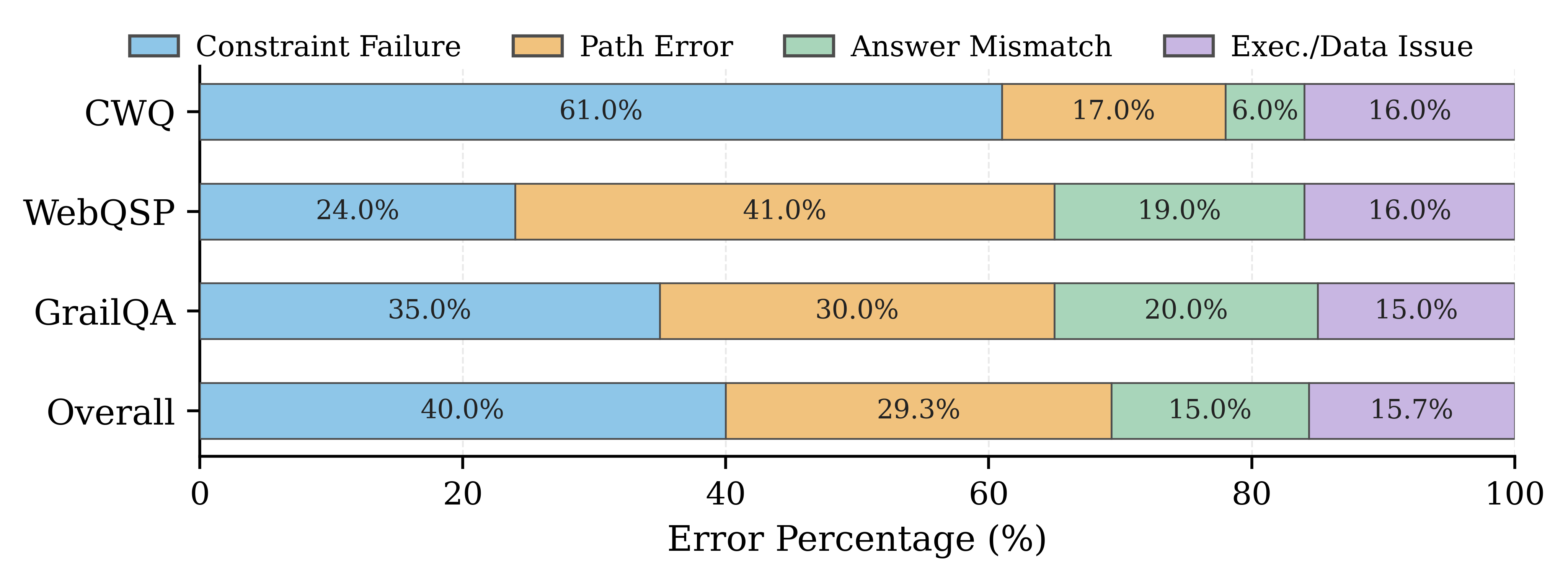}
    \caption{
Distribution of error categories among 300 randomly
sampled OACR error cases on CWQ, WebQSP, and GrailQA.
}
\label{fig:error_type_distribution}
\end{figure}

\paragraph{Hyperparameter Sensitivity.}
Figure~\ref{fig:hyperparameter_sensitivity} shows the best
overall performance at $\alpha=0.3$ and $c=0.5$.
Small $\alpha$ keeps token weights close to uniform,
while small $c$ limits the contribution of estimated
outcome-reward changes to operation credit.
Larger values can give KG-operation tokens more weight
at the expense of other policy-generated tokens.
Removing temporal ordering regularization lowers
performance, but emphasizing higher estimates at later
states can conflict with outcome-reward fitting,
potentially explaining the decline beyond $\eta=0.5$.
Increasing $\lambda$ weakens the reward for a correct
first answer, consistent with lower Hit@1.
F1 peaks at $\lambda=0.5$, suggesting that
this feedback helps predict the answer set.
\textit{These results support OACR's use of operation credit
to improve final-answer quality through different
token-level update magnitudes, rather than stronger
reallocation or temporal ordering regularization.}

\paragraph{Error Analysis.}
Figure~\ref{fig:error_type_distribution} shows that
\emph{Constraint Failure} and \emph{Path Error} account
for 69.3\% of the sampled OACR errors.
Constraint Failure dominates CWQ, whereas Path Error
is most frequent on WebQSP, indicating different
difficulties in applying question constraints and
selecting relations.
In the \emph{Answer Mismatch} case in
Appendix~\ref{app:error_case_study}, the model retrieves
the correct constellation but then expands to its
member stars and returns them as answers.
\textit{These failures suggest refining OACR's operation
credit for constraint application and relation selection,
while improving the policy's decision to end KG search
once the required evidence has been retrieved as our future work.}

\paragraph{Inference Efficiency Analysis.}
As detailed in Appendix~D.1, on CWQ and WebQSP,
OACR reduces output-token consumption by
32.3\%--72.9\% compared with KG-R1, ReknoS, and RoG
in the reported same-backbone comparisons, while
achieving higher Hit@1.
This empirical analysis excludes EoG, DPRM, MCTS-KBQA,
Graph-RFT, and SCPRM because we could not reproduce
their results using the available implementations and
settings.
However, Appendix~D.2 provides a qualitative comparison
of their inference efficiency.
\emph{To summarize, OACR combines shorter outputs with
better final-answer quality in the reported comparisons,
without invoking the KG-state outcome model during
inference.}

\section{Conclusion}
We identify a trajectory-to-operation mismatch in outcome-only GRPO
for agentic KGQA: the rollout-level update direction can reinforce
non-progressive entity-return operations and suppress progressive ones.
OACR addresses this mismatch with a KG-state outcome model that converts
estimated outcome-reward changes between adjacent KG search states into
operation credit and positive token weights.
These positive token weights rescale the outcome-level advantage
at each policy-generated token while preserving its sign.
Across three KGQA benchmarks, OACR consistently improves the
SFT-initialized policy, increases gold-answer retrieval coverage, and
reduces inference-time LLM calls and tokens. Future work will focus on maintaining question constraints across
interaction turns and using failed rollouts to improve relation selection
and relation direction.

\section*{Limitations}
We identify two limitations of the current OACR design.
First, the KG-state outcome model is periodically updated
using a bounded rollout buffer.
Its estimates may lag behind changes in the policy's KG
interactions between updates.
Adapting the update frequency could account for these
changes, but any improvement in estimation accuracy needs
to be evaluated against the additional training cost.
Second, OACR assigns one operation credit to each KG
operation, so all tokens serializing that operation share
the same weight.
This does not separately weight entity and relation
choices within an operation.
Finer-grained weighting could distinguish these choices,
but its benefit for final-answer quality remains to be
evaluated.
\bibliography{custom}

\begin{thebibliography}{36}
\providecommand{\natexlab}[1]{#1}

\bibitem[{Ao et~al.(2025)Ao, Yu, Wang, Deng, Guo, Pang, Wang, Chua, Zhang, and
  Cai}]{lightprof}
Tu~Ao, Yanhua Yu, Yuling Wang, Yang Deng, Zirui Guo, Liang Pang, Pinghui Wang,
  Tat-Seng Chua, Xiao Zhang, and Zhen Cai. 2025.
\newblock \href {https://doi.org/10.1609/aaai.v39i22.34510} {Lightprof: a
  lightweight reasoning framework for large language model on knowledge graph}.
\newblock In \emph{Proceedings of the Thirty-Ninth AAAI Conference on
  Artificial Intelligence and Thirty-Seventh Conference on Innovative
  Applications of Artificial Intelligence and Fifteenth Symposium on
  Educational Advances in Artificial Intelligence}, AAAI'25/IAAI'25/EAAI'25.
  AAAI Press.

\bibitem[{Chen et~al.(2026{\natexlab{a}})Chen, Liu, Xie, and Xiong}]{scprm}
Jiujiu Chen, Yazheng Liu, Sihong Xie, and Hui Xiong. 2026{\natexlab{a}}.
\newblock Scprm: A schema-aware cumulative process reward model for knowledge
  graph question answering.
\newblock \emph{arXiv preprint arXiv:2605.02819}.

\bibitem[{Chen et~al.(2024)Chen, Tong, Jin, Sun, Ye, and Xiong}]{pog}
Liyi Chen, Panrong Tong, Zhongming Jin, Ying Sun, Jieping Ye, and Hui Xiong.
  2024.
\newblock \href {https://doi.org/10.52202/079017-1189} {Plan-on-graph:
  Self-correcting adaptive planning of large language model on knowledge
  graphs}.
\newblock In \emph{Advances in Neural Information Processing Systems},
  volume~37, pages 37665--37691. Curran Associates, Inc.

\bibitem[{Chen et~al.(2026{\natexlab{b}})Chen, Wang, Li, Zhang, Ding, Yang, Xu,
  and Jin}]{KnowCoder-A1}
Zhuo Chen, Fei Wang, Zixuan Li, Zhao Zhang, Weiwei Ding, Chuanguang Yang,
  Yongjun Xu, and Xiaolong Jin. 2026{\natexlab{b}}.
\newblock \href {https://doi.org/10.1145/3774904.3792662} {Incentivizing
  agentic reasoning capability with outcome supervision for knowledge base
  question answering}.
\newblock In \emph{Proceedings of the ACM Web Conference 2026}, WWW '26, pages
  4302--4313, New York, NY, USA. Association for Computing Machinery.

\bibitem[{Dong et~al.(2023)Dong, Xiong, Goyal, Zhang, Chow, Pan, Diao, Zhang,
  Shum, and Zhang}]{sft3}
Hanze Dong, Wei Xiong, Deepanshu Goyal, Yihan Zhang, Winnie Chow, Rui Pan,
  Shizhe Diao, Jipeng Zhang, Kashun Shum, and Tong Zhang. 2023.
\newblock Raft: Reward ranked finetuning for generative foundation model
  alignment.
\newblock \emph{arXiv preprint arXiv:2304.06767}.

\bibitem[{Gu et~al.(2021)Gu, Kase, Vanni, Sadler, Liang, Yan, and Su}]{grailqa}
Yu~Gu, Sue Kase, Michelle Vanni, Brian Sadler, Percy Liang, Xifeng Yan, and
  Yu~Su. 2021.
\newblock Beyond iid: three levels of generalization for question answering on
  knowledge bases.
\newblock In \emph{Proceedings of the web conference 2021}, pages 3477--3488.

\bibitem[{Lightman et~al.(2024)Lightman, Kosaraju, Burda, Edwards, Baker, Lee,
  Leike, Schulman, Sutskever, and Cobbe}]{prm1}
Hunter Lightman, Vineet Kosaraju, Yuri Burda, Harrison Edwards, Bowen Baker,
  Teddy Lee, Jan Leike, John Schulman, Ilya Sutskever, and Karl Cobbe. 2024.
\newblock Let's verify step by step.
\newblock In \emph{International Conference on Learning Representations},
  volume 2024, pages 39578--39601.

\bibitem[{Liu et~al.(2025)Liu, Bai, Lu, Sun, Kong, Wang, Shan, Jose, Liu, Wen,
  Yu, and Cao}]{reweighted3}
Aiwei Liu, Haoping Bai, Zhiyun Lu, Yanchao Sun, Xiang Kong, Xiaoming~Simon
  Wang, Jiulong Shan, Albin~Madappally Jose, Xiaojiang Liu, Lijie Wen,
  Philip~S. Yu, and Meng Cao. 2025.
\newblock \href {https://openreview.net/forum?id=oF6e2WwxX0} {{TIS}-{DPO}:
  Token-level importance sampling for direct preference optimization with
  estimated weights}.
\newblock In \emph{The Thirteenth International Conference on Learning
  Representations}.

\bibitem[{Liu et~al.(2026)Liu, Feng, Qin, Chen, Yu, Gao, Yang, and
  Xu}]{prmbut2}
Yuchen Liu, Yingjie Feng, Lixiong Qin, Jiasi Chen, Jianing Yu, Sheng Gao, Sheng
  Yang, and Weiran Xu. 2026.
\newblock Beyond trajectory rewards: Step-level credit assignment for agentic
  search via graph modeling.
\newblock \emph{arXiv preprint arXiv:2605.29697}.

\bibitem[{Long et~al.(2025)Long, Zhuang, Shen, Yan, Li, and Wang}]{rtsog}
Xiao Long, Liansheng Zhuang, Chen Shen, Shaotian Yan, Yifei Li, and Shafei
  Wang. 2025.
\newblock Enhancing large language models with reward-guided tree search for
  knowledge graph question and answering.
\newblock \emph{arXiv preprint arXiv:2505.12476}.

\bibitem[{Luo et~al.(2025)Luo, E, Guo, Lin, Wu, Mu, Liu, Song, Zhu, and
  Luu}]{kbqa-o1}
Haoran Luo, Haihong E, Yikai Guo, Qika Lin, Xiaobao Wu, Xinyu Mu, Wenhao Liu,
  Meina Song, Yifan Zhu, and Anh~Tuan Luu. 2025.
\newblock \href {https://openreview.net/forum?id=QuecSemZIy} {{KBQA}-o1:
  Agentic knowledge base question answering with monte carlo tree search}.
\newblock In \emph{Forty-second International Conference on Machine Learning}.

\bibitem[{Luo et~al.(2024)Luo, Li, Haffari, and Pan}]{rog}
Linhao Luo, Yuan-Fang Li, Reza Haffari, and Shirui Pan. 2024.
\newblock Reasoning on graphs: Faithful and interpretable large language model
  reasoning.
\newblock In \emph{International Conference on Learning Representations},
  volume 2024, pages 14400--14423.

\bibitem[{Ma et~al.(2025)Ma, Chen, Wu, Khan, and Wang}]{survey1}
Chuangtao Ma, Yongrui Chen, Tianxing Wu, Arijit Khan, and Haofen Wang. 2025.
\newblock \href {https://doi.org/10.18653/v1/2025.emnlp-main.1249} {Large
  language models meet knowledge graphs for question answering: Synthesis and
  opportunities}.
\newblock In \emph{Proceedings of the 2025 Conference on Empirical Methods in
  Natural Language Processing}, pages 24578--24597, Suzhou, China. Association
  for Computational Linguistics.

\bibitem[{Nguyen et~al.(2025)Nguyen, French, Stewart, Hodkiewicz, and
  Liu}]{survey2}
Chau D.~M. Nguyen, Tim French, Michael Stewart, Melinda Hodkiewicz, and Wei
  Liu. 2025.
\newblock \href {https://doi.org/10.1145/3771692} {Representation learning in
  complex logical query answering on knowledge graphs: A survey}.
\newblock \emph{ACM Comput. Surv.}, 58(5).

\bibitem[{Ouyang et~al.(2022)Ouyang, Wu, Jiang, Almeida, Wainwright, Mishkin,
  Zhang, Agarwal, Slama, Ray, Schulman, Hilton, Kelton, Miller, Simens, Askell,
  Welinder, Christiano, Leike, and Lowe}]{sft1}
Long Ouyang, Jeffrey Wu, Xu~Jiang, Diogo Almeida, Carroll Wainwright, Pamela
  Mishkin, Chong Zhang, Sandhini Agarwal, Katarina Slama, Alex Ray, John
  Schulman, Jacob Hilton, Fraser Kelton, Luke Miller, Maddie Simens, Amanda
  Askell, Peter Welinder, Paul~F Christiano, Jan Leike, and Ryan Lowe. 2022.
\newblock \href
  {https://proceedings.neurips.cc/paper_files/paper/2022/file/b1efde53be364a73914f58805a001731-Paper-Conference.pdf}
  {Training language models to follow instructions with human feedback}.
\newblock In \emph{Advances in Neural Information Processing Systems},
  volume~35, pages 27730--27744. Curran Associates, Inc.

\bibitem[{Rezaei et~al.(2026)Rezaei, Lehmann, and Vahdati}]{prmbut3}
Mohammad Rezaei, Jens Lehmann, and Sahar Vahdati. 2026.
\newblock Llm reasoning with process rewards for outcome-guided steps.
\newblock \emph{arXiv preprint arXiv:2604.02341}.

\bibitem[{Shao et~al.(2024)Shao, Wang, Zhu, Xu, Song, Bi, Zhang, Zhang, Li, Wu
  et~al.}]{grpo}
Zhihong Shao, Peiyi Wang, Qihao Zhu, Runxin Xu, Junxiao Song, Xiao Bi, Haowei
  Zhang, Mingchuan Zhang, YK~Li, Yang Wu, et~al. 2024.
\newblock Deepseekmath: Pushing the limits of mathematical reasoning in open
  language models.
\newblock \emph{arXiv preprint arXiv:2402.03300}.

\bibitem[{Song et~al.(2025)Song, Wang, Shun, and Zhu}]{kg-r1}
Jinyeop Song, Song Wang, Julian Shun, and Yada Zhu. 2025.
\newblock Efficient and transferable agentic knowledge graph rag via
  reinforcement learning.
\newblock \emph{arXiv preprint arXiv:2509.26383}.

\bibitem[{Song et~al.(2026)Song, Liu, Guti{\'e}rrez-Basulto, Hu, Xie, Peng,
  Ananiadou, and Pan}]{Graph-RFT}
Yanlin Song, Ben Liu, V{\'\i}ctor Guti{\'e}rrez-Basulto, Zhiwei Hu, Qianqian
  Xie, Min Peng, Sophia Ananiadou, and Jeff~Z Pan. 2026.
\newblock Plan then retrieve: reinforcement learning-guided complex reasoning
  over knowledge graphs.
\newblock In \emph{Proceedings of the ACM Web Conference 2026}, pages
  3666--3676.

\bibitem[{Sun et~al.(2024)Sun, Xu, Tang, Wang, Lin, Gong, Ni, Shum, and
  Guo}]{tog}
Jiashuo Sun, Chengjin Xu, Lumingyuan Tang, Saizhuo Wang, Chen Lin, Yeyun Gong,
  Lionel Ni, Heung-Yeung Shum, and Jian Guo. 2024.
\newblock \href
  {https://proceedings.iclr.cc/paper_files/paper/2024/file/10a6bdcabbd5a3d36b760daa295f63c1-Paper-Conference.pdf}
  {Think-on-graph: Deep and responsible reasoning of large language model on
  knowledge graph}.
\newblock In \emph{International Conference on Learning Representations},
  volume 2024, pages 3868--3898.

\bibitem[{Sun et~al.(2026)Sun, Chen, Zheng, Song, Liu, Wu, Wang, Wang, and
  Wang}]{kbqa-r1}
Xin Sun, Zhongqi Chen, Xing Zheng, Bowen Song, Qiang Liu, Shu Wu, Zilei Wang,
  Weiqiang Wang, and Liang Wang. 2026.
\newblock \href {https://openreview.net/forum?id=NhWqp4R7Lf} {{KBQA}-r1:
  Reinforcing large language models for knowledge base question answering}.
\newblock In \emph{Forty-third International Conference on Machine Learning}.

\bibitem[{Talmor and Berant(2018)}]{cwq}
Alon Talmor and Jonathan Berant. 2018.
\newblock \href {https://doi.org/10.18653/v1/N18-1059} {The web as a
  knowledge-base for answering complex questions}.
\newblock In \emph{Proceedings of the 2018 Conference of the North {A}merican
  Chapter of the Association for Computational Linguistics: Human Language
  Technologies, Volume 1 (Long Papers)}, pages 641--651, New Orleans,
  Louisiana. Association for Computational Linguistics.

\bibitem[{Tian et~al.(2025)Tian, Xing, Li, Luo, Yuan, Chen, Jiang, and
  Wang}]{survey3}
Shiyu Tian, Shuyue Xing, Xingrui Li, Yangyang Luo, Caixia Yuan, Wei Chen,
  Huixing Jiang, and Xiaojie Wang. 2025.
\newblock \href {https://doi.org/10.1609/aaai.v39i24.34716} {A systematic
  exploration of knowledge graph alignment with large language models in
  retrieval augmented generation}.
\newblock In \emph{Proceedings of the Thirty-Ninth AAAI Conference on
  Artificial Intelligence and Thirty-Seventh Conference on Innovative
  Applications of Artificial Intelligence and Fifteenth Symposium on
  Educational Advances in Artificial Intelligence}, AAAI'25/IAAI'25/EAAI'25.
  AAAI Press.

\bibitem[{Wang et~al.(2026{\natexlab{a}})Wang, Zhong, Wang, Nie, Li, Yao, Li,
  and Li}]{seal}
Hao Wang, Jialun Zhong, Changcheng Wang, Zhujun Nie, Zheng Li, Shunyu Yao,
  Yanzeng Li, and Xinchi Li. 2026{\natexlab{a}}.
\newblock Seal: Self-evolving agentic learning for conversational question
  answering over knowledge graphs.
\newblock \emph{Neurocomputing}, page 133998.

\bibitem[{Wang et~al.(2024)Wang, Li, Shao, Xu, Dai, Li, Chen, Wu, and
  Sui}]{prm2}
Peiyi Wang, Lei Li, Zhihong Shao, Runxin Xu, Damai Dai, Yifei Li, Deli Chen,
  Yu~Wu, and Zhifang Sui. 2024.
\newblock Math-shepherd: Verify and reinforce llms step-by-step without human
  annotations.
\newblock In \emph{Proceedings of the 62nd Annual Meeting of the Association
  for Computational Linguistics (Volume 1: Long Papers)}, pages 9426--9439.

\bibitem[{Wang et~al.(2025)Wang, Lin, Guo, Shun, Li, and Zhu}]{ReknoS}
Song Wang, Junhong Lin, Xiaojie Guo, Julian Shun, Jundong Li, and Yada Zhu.
  2025.
\newblock \href {https://openreview.net/forum?id=rTCJ29pkuA} {Reasoning of
  large language models over knowledge graphs with super-relations}.
\newblock In \emph{The Thirteenth International Conference on Learning
  Representations}.

\bibitem[{Wang et~al.(2026{\natexlab{b}})Wang, Song, Tian, Liu, Luo, and
  Huang}]{dprm}
Xinyi Wang, Yiping Song, Zhiliang Tian, Bo~Liu, Tingjin Luo, and Minlie Huang.
  2026{\natexlab{b}}.
\newblock Dprm: A dual implicit process reward model in multi-hop question
  answering.
\newblock In \emph{Proceedings of the AAAI Conference on Artificial
  Intelligence}, volume~40, pages 33683--33691.

\bibitem[{Xiong et~al.(2025)Xiong, Li, and Zhao}]{mcts-kbqa}
Guanming Xiong, Haochen Li, and Wen Zhao. 2025.
\newblock Mcts-kbqa: Monte carlo tree search for knowledge base question
  answering.
\newblock \emph{arXiv preprint arXiv:2502.13428}.

\bibitem[{Yan et~al.(2026)Yan, Chen, Zhou, Yao, Chen, Zhang, Zhang, Zhang,
  Huang, Duan, and Zhang}]{eog}
Shiqi Yan, Yubo Chen, Ruiqi Zhou, Zhengxi Yao, Shuai Chen, Tianyi Zhang, Shijie
  Zhang, Wei-Qiang Zhang, Yongfeng Huang, Haixin Duan, and Yunqi Zhang. 2026.
\newblock \href {https://openreview.net/forum?id=NfuBj8jleE} {Explore-on-graph:
  Incentivizing autonomous exploration of large language models on knowledge
  graphs with path-refined reward modeling}.
\newblock In \emph{The Fourteenth International Conference on Learning
  Representations}.

\bibitem[{Yih et~al.(2016)Yih, Richardson, Meek, Chang, and Suh}]{webqsp}
Wen-tau Yih, Matthew Richardson, Christopher Meek, Ming-Wei Chang, and Jina
  Suh. 2016.
\newblock The value of semantic parse labeling for knowledge base question
  answering.
\newblock In \emph{Proceedings of the 54th Annual Meeting of the Association
  for Computational Linguistics (Volume 2: Short Papers)}, pages 201--206.

\bibitem[{Yuan et~al.(2025)Yuan, Li, Chen, Cui, Ding, Zhang, Zhou, Liu, and
  Peng}]{prm3}
Lifan Yuan, Wendi Li, Huayu Chen, Ganqu Cui, Ning Ding, Kaiyan Zhang, Bowen
  Zhou, Zhiyuan Liu, and Hao Peng. 2025.
\newblock Free process rewards without process labels.
\newblock In \emph{Proceedings of the 42nd International Conference on Machine
  Learning}, ICML'25. JMLR.org.

\bibitem[{Zeng et~al.(2025)Zeng, Wei, Brown, Frunza, Nevmyvaka, Zhao, and
  Hong}]{prmbut1}
Siliang Zeng, Quan Wei, William Brown, Oana Frunza, Yuriy Nevmyvaka, Yang~Katie
  Zhao, and Mingyi Hong. 2025.
\newblock Reinforcing multi-turn reasoning in llm agents via turn-level credit
  assignment.
\newblock In \emph{ICML 2025 Workshop on Computer Use Agents}.

\bibitem[{Zeng et~al.(2024)Zeng, Liu, Ma, Yang, Zhang, and Wang}]{reweighted2}
Yongcheng Zeng, Guoqing Liu, Weiyu Ma, Ning Yang, Haifeng Zhang, and Jun Wang.
  2024.
\newblock Token-level direct preference optimization.
\newblock In \emph{Proceedings of the 41st International Conference on Machine
  Learning}, ICML'24. JMLR.org.

\bibitem[{Zhang et~al.(2025)Zhang, Wen, and Zhao}]{rulekbqa}
Zhiqiang Zhang, Liqiang Wen, and Wen Zhao. 2025.
\newblock \href {https://aclanthology.org/2025.coling-main.562/} {Rule-{KBQA}:
  Rule-guided reasoning for complex knowledge base question answering with
  large language models}.
\newblock In \emph{Proceedings of the 31st International Conference on
  Computational Linguistics}, pages 8399--8417, Abu Dhabi, UAE. Association for
  Computational Linguistics.

\bibitem[{Zhou et~al.(2023)Zhou, Liu, Xu, Iyer, Sun, Mao, Ma, Efrat, Yu, YU,
  Zhang, Ghosh, Lewis, Zettlemoyer, and Levy}]{sft2}
Chunting Zhou, Pengfei Liu, Puxin Xu, Srini Iyer, Jiao Sun, Yuning Mao, Xuezhe
  Ma, Avia Efrat, Ping Yu, LILI YU, Susan Zhang, Gargi Ghosh, Mike Lewis, Luke
  Zettlemoyer, and Omer Levy. 2023.
\newblock \href {https://openreview.net/forum?id=KBMOKmX2he} {{LIMA}: Less is
  more for alignment}.
\newblock In \emph{Thirty-seventh Conference on Neural Information Processing
  Systems}.

\bibitem[{Zhou et~al.(2024)Zhou, Agrawal, Zhang, Indurthi, Zhao, Song, Xu, and
  Zhu}]{reweighted1}
Wenxuan Zhou, Ravi Agrawal, Shujian Zhang, Sathish~Reddy Indurthi, Sanqiang
  Zhao, Kaiqiang Song, Silei Xu, and Chenguang Zhu. 2024.
\newblock Wpo: Enhancing rlhf with weighted preference optimization.
\newblock In \emph{Proceedings of the 2024 Conference on Empirical Methods in
  Natural Language Processing}, pages 8328--8340.

\end{thebibliography}
\clearpage
\appendix
\setcounter{secnumdepth}{1}
\setcounter{table}{3}
\section{Related Work}
\label{app:related_work}
\paragraph{LLM-based KGQA.}
Recent work uses LLMs to retrieve facts, plan relation paths, or constrain reasoning with KG structure.
ToG~\citep{tog} performs training-free entity and relation expansion,
while RoG~\citep{rog} generates relation paths for retrieving KG facts.
PoG~\citep{pog} uses planning, memory, and reflection to revise KG
search, and ReknoS~\citep{ReknoS} uses super-relations to constrain KG
reasoning.
KBQA-o1~\citep{kbqa-o1} combines ReAct-style logical-form generation with Monte Carlo Tree Search for KG reasoning.
RTSoG~\citep{rtsog} uses reward-guided self-critic Monte Carlo Tree Search to retrieve and weight KG reasoning paths.
SEAL~\citep{seal} uses agentic calibration, memory, and reflection to refine executable S-expressions for conversational KGQA. \textit{These methods improve KGQA through inference-time search, constrained decoding, or supervised trajectories, orthogonal to our OACR that directly optimizes a multi-turn KG interaction policy from sampled rollouts.}

\paragraph{RL for agentic KGQA.}
Recent work optimizes KGQA agents from sampled rollouts.
KG-R1~\citep{kg-r1} formulates multi-turn KG interaction as a single-agent RL problem and optimizes the policy with outcome rewards.
EoG~\citep{eog} combines outcome rewards with path-refined reward modeling to guide KG exploration.
KnowCoder-A1~\citep{KnowCoder-A1} trains a ReAct-style KGQA agent through outcome-based rejection sampling and outcome-only curriculum RL.
Graph-RFT~\citep{Graph-RFT} applies reinforcement fine-tuning to complex KG reasoning with adaptive KG and web retrieval.
\textit{These methods optimize KGQA agents using rewards assigned to complete rollouts or paths.
This provides limited distinction among KG operations within the same rollout, leaving the trajectory-to-operation mismatch unresolved.
In contrast, our OACR converts estimated outcome-reward changes into operation credit to construct token weights for different KG operations.}

\paragraph{Process rewards and credit assignment.}
Prior work assigns rewards or advantages to intermediate reasoning steps or interaction turns.
PRMs score intermediate steps in multi-step LLM reasoning~\citep{prm1,prm2,prm3}, while methods for individual turns or steps estimate rewards or advantages during multi-turn agent interactions~\citep{prmbut1,prmbut2,prmbut3}.
Preference-based LLM optimization also assigns weights to training examples, text segments, or tokens~\citep{reweighted1,reweighted2,reweighted3}.
In KGQA, MCTS-KBQA~\citep{mcts-kbqa} uses a step-wise reward mechanism to guide MCTS-based logical-form construction.
DPRM~\citep{dprm} trains separate PRMs for KG and Chain-of-Thought (CoT) reasoning to derive step-level rewards from outcomes while modeling KG structural constraints and consistency between CoT reasoning and KG facts.
SCPRM~\citep{scprm} uses reasoning prefixes and schema distance to guide MCTS-based KG reasoning.
\textit{These methods provide intermediate supervision, search guidance, or optimization weights. However, accumulated intermediate rewards or advantages can change the sign of the token-level advantage. In contrast, our OACR derives weights from outcome-reward changes between adjacent KG search states and preserves the sign of the outcome-level advantage.}

\section{Dataset Statistics}
\label{dataset_stats}
We evaluate OACR on three standard KGQA benchmarks, WebQSP~\citep{webqsp}, CWQ~\citep{cwq}, and GrailQA~\citep{grailqa}. We follow the commonly used evaluation split adopted in recent work~\citep{tog, eog, KnowCoder-A1, dprm} and report results on the test split. Table~\ref{tab:testset_stats} summarizes the dataset statistics and hop distributions. The hop distribution is computed on the test split according to the KG path length required to reach the gold answer entities. WebQSP and GrailQA are dominated by one-hop questions, while CWQ contains a substantially larger fraction of two-hop and \(\geq 3\)-hop questions.

\begin{table}[!t]
\centering
\small
\setlength{\tabcolsep}{2.8pt}
\begin{adjustbox}{max width=\columnwidth}
\begin{tabular}{lrrrrrr}
\toprule
\textbf{Dataset} & \textbf{Train} & \textbf{Test} & \textbf{Max Hop} & \textbf{1-hop} & \textbf{2-hop} & \textbf{$\geq$3-hop} \\
\midrule
WebQSP  & 3,071  & 1,639 & 2 & 65.49\% & 34.51\% & 0.00\% \\
CWQ     & 27,639 & 3,519 & 4 & 40.91\% & 38.34\% & 20.75\% \\
GrailQA & 44,337 & 1,000 & 3 & 74.60\% & 24.50\% & 0.90\% \\
\bottomrule
\end{tabular}
\end{adjustbox}
\normalsize
\caption{Dataset statistics and hop distributions of questions.}
\label{tab:testset_stats}
\end{table}

Table~\ref{tab:ans_stats_test} reports their answer-set size distributions. WebQSP has a higher proportion of multi-answer questions: 51.2\% of test questions have a single answer, 27.4\% have 2--4 answers, 8.3\% have 5--9 answers, and 12.1\% have $\ge$10 answers. CWQ is dominated by single-answer questions: 70.6\% of test questions have a single answer, 19.4\% have 2--4 answers, 6.0\% have 5--9 answers, and 4.0\% have $\ge$10 answers.  GrailQA shows a distribution closer to CWQ than to WebQSP, with 68.0\% of test questions having a single answer, 18.0\% having 2--4 answers, 5.8\% having 5--9 answers, and 8.2\% having $\ge$10 answers.

\begin{table}[!t]
\centering
\small
\setlength{\tabcolsep}{3.2pt}
\begin{adjustbox}{max width=\columnwidth}
\begin{tabular*}{\columnwidth}{@{\extracolsep{\fill}} l cccc @{}}
\toprule
\textbf{Dataset} & \textbf{1} & \textbf{2--4} & \textbf{5--9} & \textbf{$\ge$10} \\
\midrule
WebQSP & 51.20\% & 27.40\% & 8.30\% & 12.10\% \\
CWQ   & 70.60\% & 19.40\% & 6.00\% & 4.00\% \\
GrailQA & 68.00\% & 18.00\% & 5.80\% & 8.20\% \\
\bottomrule
\end{tabular*}
\end{adjustbox}
\normalsize
\caption{Answer-set size distributions of test questions in WebQSP, CWQ and GrailQA. The statistics are derived from EoG~\citep{eog} and DPRM~\citep{dprm}.}
\label{tab:ans_stats_test}
\end{table}

\begin{table}[!t]
\centering
\small
\begin{tabularx}{\columnwidth}{@{}>{\raggedright\arraybackslash}Xc@{}}
\toprule
\textbf{Parameter} & \textbf{Setting} \\
\midrule
\multicolumn{2}{@{}p{\columnwidth}@{}}{\textit{\textbf{Supervised Fine-Tuning (SFT)}}} \\
\quad Batch Size & 32 \\
\quad Learning Rate & $5\times10^{-6}$ \\
\quad Weight Decay & 0.01 \\
\quad Maximum Sequence Length & 8192 \\
\quad Epochs & 3 \\
\midrule
\multicolumn{2}{@{}p{\columnwidth}@{}}{\textit{\textbf{Reinforcement Learning (RL)}}} \\
\quad Batch Size & 32 \\
\quad Rollout Group Size & 4 \\
\quad Mini-Batch Size & 16 \\
\quad Maximum Prompt Length & 2048 \\
\quad Maximum Response Length & 4096 \\
\quad Rollout Temperature & 1.0 \\
\quad Maximum Turns & 6 \\
\quad Learning Rate & $1\times10^{-6}$ \\
\quad KL Penalty Coefficient $\beta$ & 0.002 \\
\quad Epochs & 1 \\
\midrule
\multicolumn{2}{@{}p{\columnwidth}@{}}{\textit{\textbf{Outcome-Anchored Credit Reallocation (OACR)}}} \\
\quad Reallocation Strength $\alpha$ & 0.3 \\
\quad Estimated Outcome-Reward Change Bound $c$ & 0.5 \\
\quad Temporal Ordering Regularization Strength $\eta$ & 0.5 \\
\quad Outcome-Reward Mixing Coefficient $\lambda$ & 0.5 \\
\midrule
\multicolumn{2}{@{}p{\columnwidth}@{}}{\textit{\textbf{KG-State Outcome Model}}} \\
\quad Encoder & \texttt{bge-large-en-v1.5} \\
\quad Batch Size & 768 \\
\quad Learning Rate & $2\times10^{-5}$ \\
\quad Maximum Input Length & 512 \\
\quad Initial Training Epochs & 16 \\
\bottomrule
\end{tabularx}
\normalsize
\caption{Key training and rollout parameters.}
\label{tab:training_parameters}
\end{table}

\section{Implementation Details}
\label{experimental_settings}
All experiments are conducted on a server equipped with an Intel
Xeon Platinum 8470Q CPU, 128 GB RAM, and four NVIDIA RTX PRO 6000
GPUs, using Python 3.10 and PyTorch 2.8.0.
Following recent agentic KGQA methods~\citep{eog,KnowCoder-A1}, each experiment is run once with a fixed random seed of 42.
Detailed training and rollout settings are summarized in
Table~\ref{tab:training_parameters}.
We initially train the KG-state outcome model $\Phi_{\phi}$
for 16 epochs.
During OACR training, $\widetilde{\mathcal{T}}$ retains at most
8,192 complete rollouts.
Every $U=10$ RL iterations, we continue training $\Phi_{\phi}$
for one epoch using the current buffer.
Each epoch makes one pass over the distinct canonicalized
question-state pairs in the buffer, using their recomputed
targets $y(q,S)$.
Initial training and subsequent refreshes use a batch size
of 768 and a learning rate of $2\times10^{-5}$.
The update procedure is detailed in
Appendix~\ref{app:kg_state_outcome_model_details}.

\section{Inference Efficiency Analysis}
\label{inference_analysis}
\setcounter{secnumdepth}{2}

\subsection{Empirical Analysis}
\label{app:efficiency_empirical}

\begin{table*}[t]
\centering
{%
\small
\setlength{\tabcolsep}{1mm}
\begin{adjustbox}{max width=\textwidth}
\begin{tabular}{ll cccc cccc cccc}
\toprule
\multirow{2}{*}{\textbf{Method}}
& \multirow{2}{*}{\textbf{Backbone}}
& \multicolumn{4}{c}{\textbf{CWQ}}
& \multicolumn{4}{c}{\textbf{WebQSP}}
& \multicolumn{4}{c}{\textbf{GrailQA}} \\
\cmidrule(lr){3-6}
\cmidrule(lr){7-10}
\cmidrule(lr){11-14}
&
& \begin{tabular}{@{}c@{}}
    \textbf{Input} \\
    \textbf{Tokens}
  \end{tabular}
& \begin{tabular}{@{}c@{}}
    \textbf{Output} \\
    \textbf{Tokens}
  \end{tabular}
& \begin{tabular}{@{}c@{}}
    \textbf{Latency} \\
    \textbf{(s)}
  \end{tabular}
& \begin{tabular}{@{}c@{}}
    \textbf{LLM} \\
    \textbf{Calls}
  \end{tabular}
& \begin{tabular}{@{}c@{}}
    \textbf{Input} \\
    \textbf{Tokens}
  \end{tabular}
& \begin{tabular}{@{}c@{}}
    \textbf{Output} \\
    \textbf{Tokens}
  \end{tabular}
& \begin{tabular}{@{}c@{}}
    \textbf{Latency} \\
    \textbf{(s)}
  \end{tabular}
& \begin{tabular}{@{}c@{}}
    \textbf{LLM} \\
    \textbf{Calls}
  \end{tabular}
& \begin{tabular}{@{}c@{}}
    \textbf{Input} \\
    \textbf{Tokens}
  \end{tabular}
& \begin{tabular}{@{}c@{}}
    \textbf{Output} \\
    \textbf{Tokens}
  \end{tabular}
& \begin{tabular}{@{}c@{}}
    \textbf{Latency} \\
    \textbf{(s)}
  \end{tabular}
& \begin{tabular}{@{}c@{}}
    \textbf{LLM} \\
    \textbf{Calls}
  \end{tabular} \\
\midrule

ToG & \multirow{2}{*}{GPT-4}
& \underline{8182.9} & \underline{1486.4}
& \underline{96.5} & \underline{22.6}
& \underline{6031.2} & \underline{987.7}
& \underline{63.1} & \underline{15.9}
& \underline{4066.0} & \underline{774.6}
& \underline{50.2} & \underline{11.1} \\

PoG &
& \textbf{7803.0} & \textbf{353.2}
& \textbf{23.3} & \textbf{13.3}
& \textbf{5234.8} & \textbf{282.9}
& \textbf{16.8} & \textbf{9.0}
& \textbf{3372.8} & \textbf{202.8}
& \textbf{11.5} & \textbf{6.5} \\

\midrule

KG-R1 &
& \underline{2894.8} & \underline{311.1} & -- & --
& \underline{2928.5} & \underline{308.7} & -- & --
& -- & -- & -- & -- \\

\rowcolor[HTML]{DDEBFF}
\textbf{OACR} & \multirow{-2}{*}{Qwen2.5-3B}
& \textbf{2677.2} & \textbf{179.1}
& \textbf{7.2} & \textbf{3.8}
& \textbf{2006.0} & \textbf{155.2}
& \textbf{5.5} & \textbf{3.0}
& \textbf{1937.3} & \textbf{188.2}
& \textbf{7.0} & \textbf{3.3} \\

\midrule

ReknoS &
& \underline{2306.5} & 602.1 & -- & --
& 2263.9 & 558.4 & -- & 7.2
& -- & -- & -- & 10.2 \\

RoG &
& \textbf{960.3} & \underline{259.5}
& -- & \textbf{2}
& \textbf{567.7} & \underline{247.0}
& -- & \textbf{2}
& -- & -- & -- & \textbf{2} \\

KnowCoder-A1 &
& -- & -- & -- & 5.0
& -- & -- & -- & 3.9
& -- & -- & -- & -- \\

\rowcolor[HTML]{DDEBFF}
\textbf{OACR} & \multirow{-4}{*}{Qwen2.5-7B}
& 2618.5 & \textbf{175.7}
& \textbf{6.8} & \underline{3.8}
& \underline{2024.6} & \textbf{151.4}
& \textbf{6.6} & \underline{3.0}
& \textbf{1973.6} & \textbf{181.7}
& \textbf{6.7} & \underline{3.4} \\

\midrule

KBQA-o1 &
& -- & -- & -- & --
& -- & -- & -- & \underline{28.8}
& -- & -- & -- & \underline{32.3} \\

\rowcolor[HTML]{DDEBFF}
\textbf{OACR} & \multirow{-2}{*}{Llama-3.1-8B}
& \textbf{2652.7} & \textbf{176.1}
& \textbf{10.3} & \textbf{3.8}
& \textbf{2020.4} & \textbf{148.1}
& \textbf{8.6} & \textbf{3.0}
& \textbf{1993.8} & \textbf{180.1}
& \textbf{7.3} & \textbf{3.4} \\

\bottomrule
\end{tabular}
\end{adjustbox}
}
\caption{
Inference efficiency on CWQ, WebQSP, and GrailQA.
}
\label{tab:interaction_stats}
\end{table*}

Table~\ref{tab:interaction_stats} reports mean input tokens,
output tokens, LLM calls, and latency per question.
Baseline values are taken from the corresponding papers,
except for the KBQA-o1 call counts reported by
KBQA-R1~\citep{kbqa-r1}.
In the available same-backbone comparisons on CWQ and WebQSP,
OACR reduces output tokens by 32.3\%--72.9\% while achieving
higher Hit@1 (Table~\ref{tab:main_results}).
It also requires fewer LLM calls in several comparisons:
with Llama-3.1-8B on WebQSP, OACR averages 3.0 calls per
question, compared with 28.8 for the reported baseline.
These reductions do not extend uniformly to other cost
measures: the two-call Qwen2.5-7B baseline uses fewer input
and total tokens on both datasets despite generating more
output tokens.
Latency is unavailable for the same-backbone baselines,
so the reported timings do not establish a relative speedup.
\textit{The results show that OACR combines shorter outputs
with better final-answer quality in the reported comparisons,
without invoking the KG-state outcome model during inference.}

\subsection{Qualitative Analysis}
\label{app:efficiency_qualitative}

For baselines excluded from the numerical comparison, we
analyze the inference procedures described in their papers,
separating candidate generation, candidate scoring, and
retrieval.
Candidate assessments are not equivalent to LLM calls,
since multiple candidates may be processed together.
This analysis identifies sources of inference-time computation
without estimating unreported costs.

\paragraph{OACR.}
OACR generates KG operations and the final answer along
a single interaction trajectory.
The KG-state outcome model, operation-credit computation,
and token reweighting are used only during training.
Inference costs therefore arise from policy generation,
KG-tool execution, and processing accumulated observations,
without candidate scoring by $\Phi_{\phi}$.

\paragraph{EoG.}
EoG~\citep{eog} uses outcome and path-refined rewards
during RL, without a separate test-time reward evaluator
in the described procedure.
Its inference cost depends on processing the question
and KG input and generating the reasoning and answer.
The training reward design alone therefore does not
establish an inference-cost difference from OACR.

\paragraph{DPRM.}
DPRM~\citep{dprm} alternates KG-triple selection and
CoT-step generation, using KG-PRM and CoT-PRM for
Best-of-$N$ selection.
Each iteration assesses $N$ KG candidates and $N$ CoT
candidates, with $N=8$ in the paper.
Candidate generation and scoring therefore grow with
both the number of iterations and $N$, whereas OACR
requires no inference-time PRM assessments or Best-of-$N$
selection.

\paragraph{MCTS-KBQA.}
MCTS-KBQA~\citep{mcts-kbqa} expands multiple candidate
actions and uses an instruction LLM to evaluate new
intermediate states.
Its computation includes branches outside the final
selected path and depends on the expanded nodes,
candidates per expansion, and evaluation-prompt lengths.
The method bypasses terminal rollout simulation, so
branching and intermediate-state evaluation are its
main additional computations relative to OACR.

\paragraph{Graph-RFT.}
Graph-RFT~\citep{Graph-RFT} generates plans and interleaves
reasoning with relation search, neighbor search, and
Web search when KG information is insufficient.
Its inference cost includes planning, tool invocations,
and processing retrieved evidence.
Web retrieval can add latency and input tokens but may
also reduce unsuccessful KG exploration; training rewards
do not constitute test-time scoring costs.

\paragraph{SCPRM.}
SCPRM~\citep{scprm} evaluates candidate KG paths in
SCPRM-MCTS using cumulative and schema-aware future rewards.
Its scoring cost grows with the number and length of
evaluated paths.
The search returns endpoints of the highest-scoring
paths rather than generating free-form answers, so output
tokens alone omit reward-model evaluation during search.
OACR does not require this inference-time path-scoring stage.

\textit{OACR uses operation credit during training rather
than candidate scoring during inference, avoiding the
additional scoring computations of DPRM, MCTS-KBQA,
and SCPRM.
This distinction does not by itself determine latency,
which also depends on generation, retrieval, and batching.}

\section{Retrieval Coverage Analysis}
\label{app:retrieval_coverage_analysis}

\begin{figure}[!t]
    \centering
    \includegraphics[width=\columnwidth]{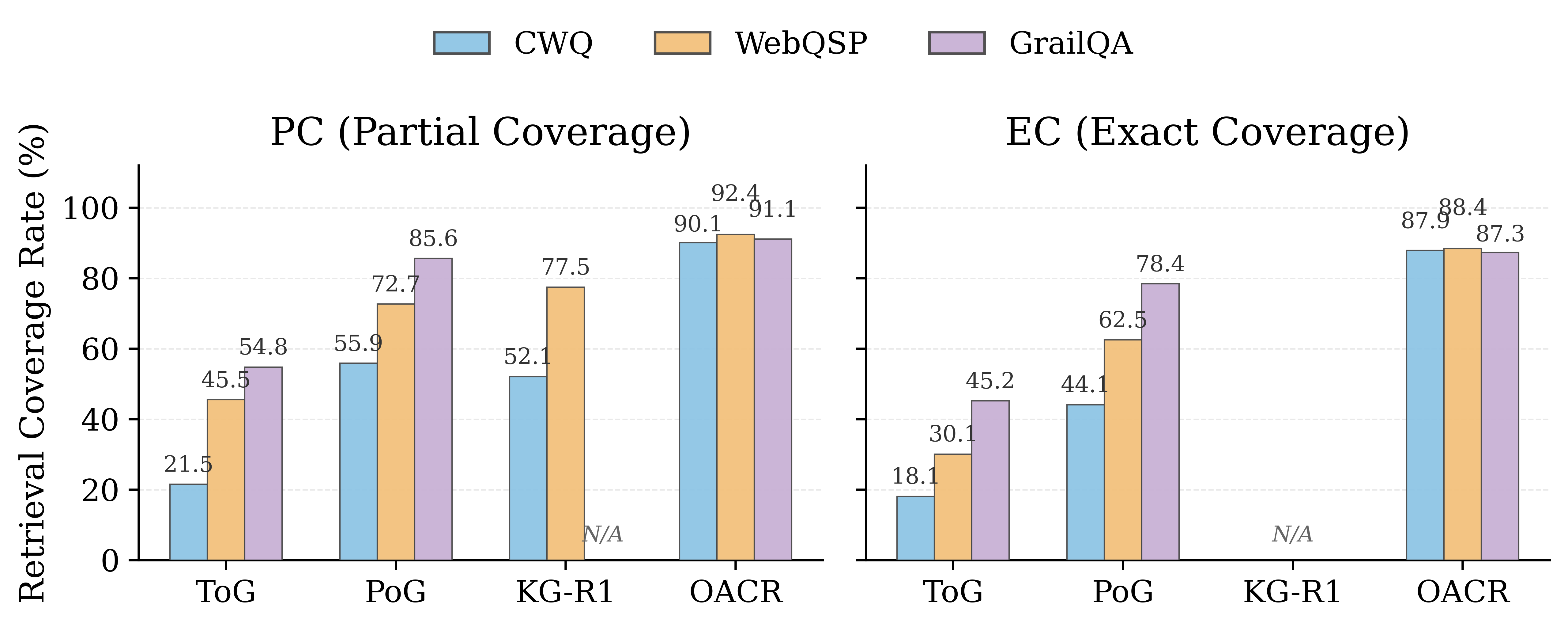}
    \caption{Retrieval coverage rate on Qwen2.5-3B.}
    \label{fig:retrieval_coverage}
\end{figure}

Figure~\ref{fig:retrieval_coverage} compares gold-answer retrieval coverage on CWQ, WebQSP, and GrailQA under the shared Qwen2.5-3B backbone.
Partial Coverage (PC) is the percentage of questions for which at least one gold answer entity is returned during KG interaction,
whereas Exact Coverage (EC) is the percentage for which the complete gold answer set is returned.
OACR achieves the highest PC and EC on all three datasets.
On CWQ, OACR reaches 90.1\% PC and 87.9\% EC, versus
55.9\% and 44.1\% for PoG.
On WebQSP, OACR achieves 92.4\% PC and 88.4\% EC, exceeding the
77.5\% PC of KG-R1 and the 62.5\% EC of PoG.
On GrailQA, OACR reaches 91.1\% PC and 87.3\% EC, versus
85.6\% and 78.4\% for PoG.
\textit{Overall, OACR retrieves gold answer entities more reliably
during KG interaction across all three datasets, showing that its
improvement is not limited to final-answer generation.}

\section{KG-State Outcome Model Details}
\label{app:kg_state_outcome_model_details}

For the KG-state outcome model $\Phi_{\phi}$, we initialize the
encoder with \texttt{bge-large-en-v1.5} and train it over serialized
question-state pairs $(q,S_t)$ with a maximum input length of 512.
Here, $S_t$ contains the ordered KG operations and returned
observations up to turn $t$.
The encoder receives only the serialized question, KG operations,
and KG observations; we do not provide manually designed structural
features or use a separate graph encoder.
When the serialized input exceeds 512 tokens, it is truncated to fit
the encoder input length, while the full canonicalized question-state
pair is retained for target aggregation in
Appendix~\ref{app:state-repeat}.

The initial rollout buffer $\widetilde{\mathcal{T}}$ is collected
using an outcome-only GRPO policy for questions
$q\sim\mathcal{Q}_{\mathrm{train}}$.
The OACR policy retains its SFT initialization $\pi_{\theta_0}$.
After each OACR policy update, we add the sampled rollouts
and their outcome rewards to $\widetilde{\mathcal{T}}$.
The buffer has a fixed capacity and retains the most recent rollouts;
the initial GRPO rollouts are gradually replaced by OACR rollouts.
Rollouts are retained regardless of their outcome rewards.

Every $U$ RL iterations, after the policy update, we recompute
$y(q,S)$ from the rollouts currently retained in
$\widetilde{\mathcal{T}}$ and continue training the existing
$\Phi_{\phi}$ with Eq.~\ref{eq:kg_state_outcome_loss}.
The updated model scores subsequent newly sampled rollouts.
During each policy update, $\Phi_{\phi}$ and the cached
outcome-reward estimates and token weights are held fixed,
and the policy loss does not propagate gradients through them.
$\Phi_{\phi}$ is not used during inference.

Algorithm~\ref{alg:kg_state_outcome_model} fits the KG-state outcome
model to a given rollout buffer.
Algorithm~\ref{alg:oacr} specifies how this fitting procedure is
used before and during OACR policy optimization.

\begin{algorithm*}[t]
\caption{Training the KG-State Outcome Model}
\label{alg:kg_state_outcome_model}
\begin{algorithmic}[1]
\REQUIRE Initialized KG-state outcome model $\Phi_{\phi}$,
rollout buffer $\widetilde{\mathcal{T}}$,
$\lambda$, $\eta$, optimization budget
\ENSURE Updated KG-state outcome model $\Phi_{\phi}$
\FOR{each $\widetilde{\tau}_m\in\widetilde{\mathcal{T}}$}
    \STATE Compute its outcome reward $\widetilde{R}_m$.
    \hfill$\triangleright$ Eq.~\ref{eq:outcome_reward}
    \FOR{$t=1,\ldots,L_m$}
        \STATE Canonicalize $(q_m,\widetilde{S}_{m,t})$ as
        $(q,S)$ and record $(q,S,\widetilde{R}_m)$.
    \ENDFOR
\ENDFOR
\STATE Recompute $y(q,S)$ by averaging outcome rewards
for repeated $(q,S)$ in the current buffer.
\STATE Compute the fitting loss $\mathcal{L}_{\mathrm{fit}}$.
\hfill$\triangleright$ Eq.~\ref{eq:kg_state_fit_loss}
\STATE Compute the temporal ordering regularizer
$\mathcal{L}_{\mathrm{ord}}$.
\hfill$\triangleright$ Eq.~\ref{eq:kg_state_order_loss}
\STATE Optimize $\phi$ with $\mathcal{L}_{\Phi}$
for the specified budget.
\hfill$\triangleright$ Eq.~\ref{eq:kg_state_outcome_loss}
\RETURN $\Phi_{\phi}$.
\end{algorithmic}
\end{algorithm*}

\begin{algorithm*}[t]
\caption{Outcome-Anchored Credit Reallocation (OACR)}
\label{alg:oacr}
\begin{algorithmic}[1]
\REQUIRE SFT training set $\mathcal{D}$,
knowledge graph $\mathcal{G}$,
RL question distribution $\mathcal{Q}_{\mathrm{train}}$,
fixed reference policy $\pi_{\mathrm{ref}}$,
rollout group size $B$, RL iterations $K$,
initial outcome-only GRPO rollout buffer
$\widetilde{\mathcal{T}}$,
refresh interval $U$, buffer capacity,
initial and refresh optimization budgets for $\Phi_{\phi}$
\ENSURE Optimized KGQA policy $\pi_{\theta}$
\STATE Train the SFT-initialized policy $\pi_{\theta_0}$
on $\mathcal{D}$.
\hfill$\triangleright$ Eq.~\ref{eq:sft_loss}
\STATE Initialize $\Phi_{\phi}$ and fit it to
$\widetilde{\mathcal{T}}$ with
Algorithm~\ref{alg:kg_state_outcome_model}.
\STATE Initialize $\theta\leftarrow\theta_0$.
\FOR{$k=1,\ldots,K$}
    \STATE Set $\theta_{\mathrm{old}}\leftarrow\theta$ and sample $q\sim\mathcal{Q}_{\mathrm{train}}$ with rollout group $\{\tau_i\}_{i=1}^{B}$. \hfill$\triangleright$ Eq.~(4)
    \STATE Compute each outcome reward $R_i$ and outcome-level advantage $A_i$. \hfill$\triangleright$ Eqs.~(5)--(7)
    \FOR{each rollout $\tau_i$}
        \STATE Cache $\Phi_{\phi}(q,S_{i,t})$ for all visited KG search states.
        \STATE Compute estimated outcome-reward changes $d_{i,t}$. \hfill$\triangleright$ Eq.~(14)
        \STATE Convert $d_{i,t}$ into operation credit $b_{i,t}$ conditioned on $A_i$. \hfill$\triangleright$ Eq.~(15)
        \STATE Construct raw token weights $\omega_{i,j}$ and normalize them to $w_{i,j}$. \hfill$\triangleright$ Eqs.~(16)--(17)
        \STATE Compute the reallocated token-level advantage $\widetilde{A}_{i,j}$. \hfill$\triangleright$ Eq.~(18)
    \ENDFOR
        \STATE Update $\theta$ using the OACR objective,
    holding $\Phi_{\phi}$ and cached quantities fixed.
    \hfill$\triangleright$ Eq.~\ref{eq:oacr_loss}
    \STATE Add the sampled rollouts and their outcome rewards
    to $\widetilde{\mathcal{T}}$; discard the oldest rollouts
    if its capacity is exceeded.
    \IF{$k\bmod U=0$}
        \STATE Continue fitting $\Phi_{\phi}$ to the current
        $\widetilde{\mathcal{T}}$ with
        Algorithm~\ref{alg:kg_state_outcome_model}.
    \ENDIF
\ENDFOR
\RETURN $\pi_{\theta}$
\end{algorithmic}
\end{algorithm*}

\section{State-Visit Statistics and Estimated Outcome-Reward Change Analysis}
\label{app:state-repeat}

We analyze repeated visits to the canonicalized question-state
pairs used to train the KG-state outcome model.
For each buffered rollout, every intermediate KG search state
$S_t$ contributes one visit to the corresponding question-state
pair $(q,S_t)$, while the initial empty state $S_0$ is excluded.

Before aggregating visits, we canonicalize each question-state
pair. For each relation-list operation, the \texttt{action} and
\texttt{entity} fields are serialized in a fixed order.
For each entity-return operation, the \texttt{action},
\texttt{entity}, and \texttt{relation} fields are serialized in
a fixed order.
Duplicate strings in each returned relation or entity list
are removed, and the remaining strings are sorted
lexicographically before the returned observation is serialized.
The order of KG operations and returned observations across
turns is preserved.

Consequently, two question-state pairs are assigned the same
aggregation key only when they contain the same question and
the same ordered sequence of KG operations and canonicalized
returned observations.
This canonicalization removes differences caused only by field
order, duplicate returned items, or the order of items within
a returned list, without changing the KG search state across
turns.
The full canonicalized question-state pair is used as the
aggregation key without observation truncation or lossy hashing.
The 512-token maximum input length is applied only to the
encoder input and does not affect target aggregation.

Therefore, repeated question-state pairs in
Table~\ref{tab:state-repeat} correspond to exact canonical
repeats under these rules rather than collisions caused by
truncation or hashing.

\begin{table*}[t]
\centering
\small
\begin{adjustbox}{max width=\textwidth}
\begin{tabular}{lrrrrrr}
\toprule
Subset & \#Visits & \#Pairs & Mean & Median
& Singleton (\%) & Visit-wtd. $\geq 2$ (\%) \\
\midrule
All      & 120,144 & 33,887 & 3.55 & 4 & 23.37 & 93.41 \\
CWQ      & 55,291  & 16,396 & 3.37 & 4 & 25.57 & 92.42 \\
WebQSP   & 4,373   & 1,251  & 3.50 & 4 & 25.02 & 92.84 \\
GrailQA  & 60,480  & 16,240 & 3.72 & 4 & 21.02 & 94.36 \\
Depth 1  & 88,826  & 22,858 & 3.89 & 4 & 14.65 & 96.23 \\
Depth 2  & 29,126  & 9,096  & 3.20 & 4 & 31.04 & 90.31 \\
Depth 3+ & 2,192   & 1,933  & 1.13 & 1 & 90.27 & 20.39 \\
\bottomrule
\end{tabular}
\end{adjustbox}
\normalsize
\caption{
Visit-count statistics of canonicalized question-state pairs
in the buffered rollouts used to train $\Phi_{\phi}$.
The initial empty state $S_0$ is excluded.
``Singleton'' is the percentage of unique question-state pairs
visited once.
``Visit-wtd. $\geq 2$'' is the percentage of visits contributed
by question-state pairs visited at least twice.
}
\label{tab:state-repeat}
\end{table*}

As shown in Table~\ref{tab:state-repeat}, the buffer contains
120,144 intermediate-state visits and 33,887 distinct
canonicalized question-state pairs.
Overall, 76.63\% of unique question-state pairs are visited
at least twice, and these repeated pairs account for
93.41\% of all visits.
At Depth 1 and Depth 2, repeated pairs contribute
96.23\% and 90.31\% of visits, respectively, indicating
that early KG search states are frequently revisited.
In contrast, at Depth 3 or deeper, 90.27\% of unique
question-state pairs are singletons, and only 20.39\%
of visits belong to repeated pairs.
However, these deeper states account for only 2,192 of
120,144 visits (1.82\%).
\textit{Overall, most buffered state visits belong to
repeated canonicalized question-state pairs, particularly
at early search depths, while deeper states are substantially
sparser.}

We further analyze whether the fitted KG-state outcome model
assigns different estimated outcome-reward changes to
progressive and non-progressive entity-return operations.
To make this diagnostic consistent with
Figure~\ref{fig:outcome_credit_issue}, we use the same CWQ
diagnostic rollouts generated by the Qwen2.5-3B
outcome-only GRPO model.
For each entity-return operation, we compute $d_{i,t}$
using Eq.~\ref{eq:estimated_outcome_reward_change} and
assign the distance-based progressive/non-progressive label
following Figure~\ref{fig:outcome_credit_issue}.

For successful and failed rollouts separately,
$\mathrm{Avg.}\,d_{\mathrm{prog}}$ and
$\mathrm{Avg.}\,d_{\mathrm{non}}$ denote the mean clipped
estimated outcome-reward changes over progressive and
non-progressive entity-return operations, respectively.
Gap denotes their difference,
$\mathrm{Avg.}\,d_{\mathrm{prog}}-
\mathrm{Avg.}\,d_{\mathrm{non}}$.

\begin{table*}[t]
\centering
\begin{adjustbox}{max width=\textwidth}
\begin{tabular}{lrrrrr}
\toprule
Group & progressive & non-progressive
& Avg. $d_{\mathrm{prog}}$
& Avg. $d_{\mathrm{non}}$ & Gap \\
\midrule
Successful rollouts & 34.2\% & 65.8\% & 0.24 & -0.17 & 0.41 \\
Failed rollouts     & 25.3\% & 74.7\% & 0.21 & -0.26 & 0.47 \\
\bottomrule
\end{tabular}
\end{adjustbox}
\normalsize
\caption{
Alignment between estimated outcome-reward changes and
distance-based progressive/non-progressive labels on the
same CWQ diagnostic rollouts used in
Figure~\ref{fig:outcome_credit_issue}.
``Progressive'' and ``non-progressive'' follow the
distance-based labels in Figure~\ref{fig:outcome_credit_issue}.
}
\label{tab:phi-change-gap}
\end{table*}

Table~\ref{tab:phi-change-gap} reports the average estimated
outcome-reward changes for progressive and non-progressive
operations in successful and failed rollouts.
The class distribution follows
Figure~\ref{fig:outcome_credit_issue}: non-progressive
operations remain frequent in successful rollouts, while
progressive operations also occur in failed rollouts.
In both groups, progressive operations receive larger
estimated outcome-reward changes than non-progressive
operations, with gaps of 0.41 and 0.47, respectively.
\textit{This alignment with the separately defined
distance-based labels supports using estimated
outcome-reward changes from $\Phi_{\phi}$ to reweight,
rather than replace, the outcome-level advantage.}

\section{SFT Trajectory Examples}
\label{app:sft_examples}

Tables~\ref{tab:sft-cwq}, \ref{tab:sft-webqsp}, and
\ref{tab:sft-grailqa} provide representative SFT trajectories on
CWQ, WebQSP, and GrailQA, respectively.
Each trajectory follows the serialized interaction format used during
inference: the policy generates natural-language reasoning in
$<$think$>$ blocks, issues schema-valid KG
operations in $<$search$>$ blocks, receives returned
KG observations in $<$observation$>$ blocks, and
emits the final answer in $<$answer$>$ blocks.
These examples show how SFT initializes the policy to follow the
KG-tool schema, select relation directions, copy entity and relation
strings from returned KG observations, continue generation conditioned
on those observations, and follow the final-answer format.
For readability, long relation and entity lists in the returned KG
observations are truncated with ellipses.

\section{Inference Rollout Examples}
\label{app:oacr_inference_examples}

To illustrate how the OACR policy issues KG operations during
inference, we provide representative successful rollouts from the
evaluation sets in Tables~\ref{tab:oacr_infer_cwq},
\ref{tab:oacr_infer_webqsp}, and
\ref{tab:oacr_infer_grailqa}.
Each example reports the question, gold answer set, predicted answer
list, issued KG operations, and returned KG observations.
For readability, long returned KG observations are truncated with
ellipses.

\section{Detailed Error Case Analysis}
\label{app:error_case_study}

To illustrate the remaining failure modes of OACR, we present three
representative failed rollouts in
Tables~\ref{tab:case_constraint_failure},
\ref{tab:case_path_error}, and
\ref{tab:case_answer_mismatch}.
These rollouts correspond to the major error categories discussed in
the main-paper error analysis: Constraint Failure, Path Error, and
Answer Mismatch.

\section{Detailed Results on GrailQA Generalization Splits}
\label{app:grailqa_split_results}
Table~\ref{tab:grailqa_split_results} further reports F1 and Hit@1
on the I.I.D., compositional, and zero-shot splits for methods in
Table 1 with available split-wise results.
Baselines without reported GrailQA split-wise results are omitted.
Across all three backbones, OACR improves over
OACR$_{\mathrm{SFT}}$ on both F1 and Hit@1 for the I.I.D.,
compositional, and zero-shot splits.
The gains are relatively balanced under Qwen2.5-3B, whereas the two
stronger backbones show their largest improvements on the
compositional split: 15.6/13.4 F1/Hit@1 points with Qwen2.5-7B and
14.8/11.6 points with Llama-3.1-8B.
OACR also substantially improves the overall GrailQA results under
each backbone.
\textit{Overall, OACR consistently improves the SFT-initialized
policy across all GrailQA generalization splits, with particularly
large gains on the compositional split under the stronger backbones,
showing that its improvements do not arise from only one
generalization setting.}
% Case table style
% =========================

% ----- text commands -----
\newcommand{\casefield}[1]{\textbf{#1}}
\newcommand{\casecode}[1]{\texttt{\small #1}}
\newcommand{\casegold}[1]{\textcolor{GoldBlue}{#1}}
\newcommand{\casepred}[1]{\textcolor{PredRed}{#1}}

\begin{table*}
\centering

\small
\setlength{\tabcolsep}{7pt}

\begin{adjustbox}{max width=\textwidth}
\begin{tabular}{
  >{\raggedright\arraybackslash}p{0.16\textwidth}
  >{\raggedright\arraybackslash}p{0.78\textwidth}
}
\hline

\rowcolor{PromptBg}
\casefield{System Content}
&
You are a KGQA agent. Reason in \textcolor{violet}{$<$think$>$...$</$think$>$}, issue KG operations in \textcolor{cyan}{$<$search$>$...$</$search$>$}, and give the final answer in \textcolor{magenta}{$<$answer$>$...$</$answer$>$}.

Available KG operations:

1. \{``action'':``get\_tail\_relations'',``entity'':``$<$entity\_name$>$''\}

2. \{``action'':``get\_tail\_entities'',``entity'':``$<$entity\_name$>$'',``relation'':``$<$relation\_id$>$''\}

3. \{``action'':``get\_head\_relations'',``entity'':``$<$entity\_name$>$''\}

4. \{``action'':``get\_head\_entities'',``entity'':``$<$entity\_name$>$'',``relation'':``$<$relation\_id$>$''\}

Final answer format: \textcolor{magenta}{$<$answer$>$}\{``answers'':[``name1'',``name2'']\}\textcolor{magenta}{$<$/answer$>$}

\\
\midrule

\rowcolor{PromptBg}
\casefield{User Content}
&
Question: What are the titles of the members of the office of William R. King?

Start entities: \textcolor{orange}{William R. King}
\\
\hline

\rowcolor{TurnBg}
\casefield{Assistant Content}
&
\textcolor{violet}{$<$think$>$}
I need to inspect incoming relations for \textcolor{orange}{``William R. King''}.
\textcolor{violet}{$<$/think$>$}

\textcolor{cyan}{$<$search$>$}\{``action'':``get\_head\_relations'',``entity'':``\textcolor{orange}{William R. King}''\}\textcolor{cyan}{$<$/search$>$}
\\
\midrule

\rowcolor{TurnBg}
\casefield{User Content}
&
\textcolor{gray}{$<$observation$>$}
\seqsplit{\{``relations'':[``government.government\_position\_held.office\_holder'',``government.political\_party\_tenure.politician'',``government.us\_president.vice\_president'',``people.place\_lived.person'',``book.written\_work.author'', ...]\}}
\textcolor{gray}{$<$/observation$>$}

Based on the observation, continue reasoning in \textcolor{violet}{$<$think$>$} tags and issue your next KG operation in \textcolor{cyan}{$<$search$>$} tags, or provide the final answer in \textcolor{magenta}{$<$answer$>$} tags.
\\
\midrule

\rowcolor{TurnBg}
\casefield{Assistant Content}
&
\textcolor{violet}{$<$think$>$}
The relation ``\seqsplit{government.government\_position\_held.office\_holder}'' connects government-position records to their office holder. I will follow it from \textcolor{orange}{``William R. King''} to obtain the related office-title information.
\textcolor{violet}{$<$/think$>$}

\textcolor{cyan}{$<$search$>$}\{``action'':``get\_head\_entities'',``entity'':``\textcolor{orange}{William R. King}'',``relation'':``\seqsplit{government.government\_position\_held.office\_holder}''\}\textcolor{cyan}{$<$/search$>$}
\\
\midrule

\rowcolor{TurnBg}
\casefield{User Content}
&
\textcolor{gray}{$<$observation$>$}
\{``entities'':[``Alabama'',``United States Senate'',\textcolor{blue}{``Senator''},\textcolor{blue}{``Chief of staff''},\textcolor{blue}{``President of the Senate''},\textcolor{blue}{``State Senator''},``United States of America'',``United States Senator'',``26th United States Congress'',``27th United States Congress'',``28th United States Congress'', ...]\}
\textcolor{gray}{$<$/observation$>$}

Based on the observation, continue reasoning in \textcolor{violet}{$<$think$>$} tags and issue your next KG operation in \textcolor{cyan}{$<$search$>$} tags, or provide the final answer in \textcolor{magenta}{$<$answer$>$} tags.
\\
\midrule

\rowcolor{TurnBg}
\casefield{Assistant Content}
&
\textcolor{violet}{$<$think$>$}
The latest entity observation contains the required office-title answers, so I can answer directly.
\textcolor{violet}{$<$/think$>$}

\textcolor{magenta}{$<$answer$>$}\{``answers'':[``Senator'',``Chief of staff'',``President of the Senate'',``State Senator'']\}\textcolor{magenta}{$<$/answer$>$}
\\
\hline

\rowcolor{AnswerBg}
\casefield{Gold Answer}
&
\textcolor{red}{``Senator'',``Chief of staff'',``President of the Senate'',``State Senator''}
\\
\hline

\end{tabular}
\end{adjustbox}
\normalsize
\caption{SFT trajectory example on CWQ.}
\label{tab:sft-cwq}
\end{table*}

\begin{table*}
\centering

\small
\setlength{\tabcolsep}{7pt}

\begin{adjustbox}{max width=\textwidth}
\begin{tabular}{
  >{\raggedright\arraybackslash}p{0.16\textwidth}
  >{\raggedright\arraybackslash}p{0.78\textwidth}
}
\hline

\rowcolor{PromptBg}
\casefield{System Content}
&
You are a KGQA agent. Reason in \textcolor{violet}{$<$think$>$...$</$think$>$}, issue KG operations in \textcolor{cyan}{$<$search$>$...$</$search$>$}, and give the final answer in \textcolor{magenta}{$<$answer$>$...$</$answer$>$}.

Available KG operations:

1. \{``action'':``get\_tail\_relations'',``entity'':``$<$entity\_name$>$''\}

2. \{``action'':``get\_tail\_entities'',``entity'':``$<$entity\_name$>$'',``relation'':``$<$relation\_id$>$''\}

3. \{``action'':``get\_head\_relations'',``entity'':``$<$entity\_name$>$''\}

4. \{``action'':``get\_head\_entities'',``entity'':``$<$entity\_name$>$'',``relation'':``$<$relation\_id$>$''\}

Final answer format: \textcolor{magenta}{$<$answer$>$}\{``answers'':[``name1'',``name2'']\}\textcolor{magenta}{$<$/answer$>$}

\\
\midrule

\rowcolor{PromptBg}
\casefield{User Content}
&
Question: who is the senator of the united states 2010?

Start entities: \textcolor{orange}{United States of America}
\\
\hline

\rowcolor{TurnBg}
\casefield{Assistant Content}
&
\textcolor{violet}{$<$think$>$}
I need to inspect outgoing relations for \textcolor{orange}{``United States of America''}.
\textcolor{violet}{$<$/think$>$}

\textcolor{cyan}{$<$search$>$}\{``action'':``get\_tail\_relations'',``entity'':``\textcolor{orange}{United States of America}''\}
\textcolor{cyan}{$<$/search$>$}
\\
\midrule

\rowcolor{TurnBg}
\casefield{User Content}
&
\textcolor{gray}{$<$observation$>$}
\{``relations'':[``\seqsplit{military.military\_combatant.military\_commanders}'',``\seqsplit{government.political\_district.representatives}'',``\seqsplit{location.location.people\_born\_here}'',``\seqsplit{location.country.fips10\_4}'',``\seqsplit{government.governmental\_jurisdiction.governing\_officials}'',``\seqsplit{military.military\_combatant.belongs\_to\_combatant\_group}'',``\seqsplit{olympics.olympic\_participating\_country.athletes}'',``\seqsplit{military.military\_combatant.casualties}'',``\seqsplit{military.military\_combatant.armed\_forces}'',``\seqsplit{location.location.events}'', ...]\}
\textcolor{gray}{$<$/observation$>$}

Based on the observation, continue reasoning in \textcolor{violet}{$<$think$>$} tags and issue your next KG operation in \textcolor{cyan}{$<$search$>$} tags, or provide the final answer in \textcolor{magenta}{$<$answer$>$} tags.
\\
\midrule

\rowcolor{TurnBg}
\casefield{Assistant Content}
&
\textcolor{violet}{$<$think$>$}The relation ``\seqsplit{government.governmental\_jurisdiction.governing\_officials}'' is relevant, so I will follow it from \textcolor{orange}{``United States of America''}.\textcolor{violet}{$<$/think$>$}

\textcolor{cyan}{$<$search$>$}\{``action'':``get\_tail\_entities'',``entity'':``\textcolor{orange}{United States of America}'',``relation'':``\seqsplit{government.governmental\_jurisdiction.governing\_officials}''\}\textcolor{cyan}{$<$/search$>$}
\\
\midrule

\rowcolor{TurnBg}
\casefield{User Content}
&
\textcolor{gray}{$<$observation$>$}
\{``entities'':[``New York'',``United States Senate'',``Edward Murphy, Jr.'',``Senator'',``1893-03-04'',``United States Senator'',``1899-03-03'',``53rd United States Congress'',``55th United States Congress'',``54th United States Congress'',``Tennessee'',``Felix Grundy'',``1839-12-14'',``1840-12-19'',``26th United States Congress'',``South Carolina'',``Robert Rhett'',``1850-12-18'',``1852-05-07'',``32nd United States Congress'',``31st United States Congress'',``California'',``James D. Phelan'',``1915-03-04'',``1921-03-04'',``Alabama'',``Henry H. Chambers'',``1825-03-04'',``1826-01-24'',``19th United States Congress'', ...]\}
\textcolor{gray}{$<$/observation$>$}

Based on the observation, continue reasoning in \textcolor{violet}{$<$think$>$} tags and issue your next KG operation in \textcolor{cyan}{$<$search$>$} tags, or provide the final answer in \textcolor{magenta}{$<$answer$>$} tags.
\\
\midrule

\rowcolor{TurnBg}
\casefield{Assistant Content}
&
\textcolor{violet}{$<$think$>$}The latest entity observation contains the answer name(s), so I can answer directly.\textcolor{violet}{$<$/think$>$}

\textcolor{magenta}{$<$answer$>$}\{``answers'':[``Jeanne Shaheen'',``Daniel Inouye'',``Robert Byrd'',``Harry Reid'',``John Barrasso'',``Jon Kyl'',``John McCain'',``Blanche Lincoln'',``Bernie Sanders'',``Tim Johnson'',``Kirsten Gillibrand'']\}\textcolor{magenta}{$<$/answer$>$}
\\
\hline

\rowcolor{AnswerBg}
\casefield{Gold Answer}
&
\textcolor{red}{``Jeanne Shaheen'',``Daniel Inouye'',``Robert Byrd'',``Jefferson Sessions'',``Mark Warner'',``Harry Reid'',``Russ Feingold'',``Tom Udall'',``Jeff Bingaman'',``Sheldon Whitehouse'',``Lisa Murkowski'',``Claire McCaskill'',``Scott Brown'',``Chuck Schumer'',``Tom Harkin'', ...}

\\
\hline
\end{tabular}
\end{adjustbox}
\normalsize
\caption{SFT trajectory example on WebQSP.}
\label{tab:sft-webqsp}
\end{table*}

\begin{table*}
\centering

\small
\setlength{\tabcolsep}{7pt}

\begin{adjustbox}{max width=\textwidth}
\begin{tabular}{
  >{\raggedright\arraybackslash}p{0.16\textwidth}
  >{\raggedright\arraybackslash}p{0.78\textwidth}
}
\hline

\rowcolor{PromptBg}
\casefield{System Content}
&
You are a KGQA agent. Reason in \textcolor{violet}{$<$think$>$...$</$think$>$}, issue KG operations in \textcolor{cyan}{$<$search$>$...$</$search$>$}, and give the final answer in \textcolor{magenta}{$<$answer$>$...$</$answer$>$}.

Available KG operations:

1. \{``action'':``get\_tail\_relations'',``entity'':``$<$entity\_name$>$''\}

2. \{``action'':``get\_tail\_entities'',``entity'':``$<$entity\_name$>$'',``relation'':``$<$relation\_id$>$''\}

3. \{``action'':``get\_head\_relations'',``entity'':``$<$entity\_name$>$''\}

4. \{``action'':``get\_head\_entities'',``entity'':``$<$entity\_name$>$'',``relation'':``$<$relation\_id$>$''\}

Final answer format: \textcolor{magenta}{$<$answer$>$}\{``answers'':[``name1'',``name2'']\}\textcolor{magenta}{$<$/answer$>$}

\\
\midrule

\rowcolor{PromptBg}
\casefield{User Content}
&
Question: what venture funded company has chief executive in its organization leadership?

Start entities: \textcolor{orange}{chief executive}
\\
\hline

\rowcolor{TurnBg}
\casefield{Assistant Content}
&
\textcolor{violet}{$<$think$>$}
I need to inspect incoming relations for \textcolor{orange}{``chief executive''}.
\textcolor{violet}{$<$/think$>$}

\textcolor{cyan}{$<$search$>$}\{``action'':``get\_head\_relations'',``entity'':``\textcolor{orange}{chief executive}''\}
\textcolor{cyan}{$<$/search$>$}
\\
\midrule

\rowcolor{TurnBg}
\casefield{User Content}
&
\textcolor{gray}{$<$observation$>$}
\{``relations'':[``\seqsplit{organization.leadership.role}'',``\seqsplit{organization.organization.founders}'',``\seqsplit{organization.organization\_board\_membership.role}'',``\seqsplit{organization.email\_contact.category}'',``\seqsplit{business.employment\_tenure.company}'',``\seqsplit{business.employment\_tenure.title}'',``\seqsplit{projects.project\_participation.role}'',``\seqsplit{fictional\_universe.fictional\_employment\_tenure.title}'',``\seqsplit{people.person.profession}'',``\seqsplit{fictional\_universe.fictional\_character.rank}'',``\seqsplit{fictional\_universe.fictional\_character.occupation}'',``\seqsplit{book.written\_work.subjects}'',``\seqsplit{measurement\_unit.dated\_integer.source}'']\}
\textcolor{gray}{$<$/observation$>$}

Based on the observation, continue reasoning in \textcolor{violet}{$<$think$>$} tags and issue your next KG operation in \textcolor{cyan}{$<$search$>$} tags, or provide the final answer in \textcolor{magenta}{$<$answer$>$} tags.
\\
\midrule

\rowcolor{TurnBg}
\casefield{Assistant Content}
&
\textcolor{violet}{$<$think$>$}The relation ``\seqsplit{organization.leadership.role}'' is relevant, so I will follow it from \textcolor{orange}{``chief executive''}.\textcolor{violet}{$<$/think$>$}

\textcolor{cyan}{$<$search$>$}\{``action'':``get\_head\_entities'',``entity'':``\textcolor{orange}{chief executive}'',``relation'':``\seqsplit{organization.leadership.role}''\}\textcolor{cyan}{$<$/search$>$}
\\
\midrule

\rowcolor{TurnBg}
\casefield{User Content}
&
\textcolor{gray}{$<$observation$>$}
\{``entities'':[``2014-07-17'',``Mothercare'',``Mark Newton-Jones'',``Chief Executive Officer'',``2009'',``Embarq Corporation'',``Glen F. Post, III'',``President \& CEO'',``OneLink Communications'',``Ron Dorchester'',``2000'',``Scripps Health'',``Chris Van Gorder'',``President and CEO'',``2011-09-01'',``Sony Computer Entertainment International'',``Andrew House'',``President and Chief Executive Officer'',``2006-06'',``Spirit AeroSystems'',``Jeffrey L Turner'',``2013-04'',``2011-02-28'',``Veda'',``Nerida Caesar'',``2012-05-01'',``Suncor Energy'',``Steve Williams'',``2007-08-06'',``CEVA Logistics'', ...]\}
\textcolor{gray}{$<$/observation$>$}

Based on the observation, continue reasoning in \textcolor{violet}{$<$think$>$} tags and issue your next KG operation in \textcolor{cyan}{$<$search$>$} tags, or provide the final answer in \textcolor{magenta}{$<$answer$>$} tags.
\\
\midrule

\rowcolor{TurnBg}
\casefield{Assistant Content}
&
\textcolor{violet}{$<$think$>$}The latest entity observation contains the answer name(s), so I can answer directly.\textcolor{violet}{$<$/think$>$}

\textcolor{magenta}{$<$answer$>$}\{``answers'':[``Nodality'',``Guidewire Software'',``Amec Foster Wheeler'',``Aricent'',``LinkedIn Corporation'',``Audible Inc.'',``StumbleUpon'',``Comcast'',``Synaptics'',``ZoomInfo'',``Nokia'',``eBags.com'',``StrongMail Systems'']\}\textcolor{magenta}{$<$/answer$>$}
\\
\hline

\rowcolor{AnswerBg}
\casefield{Gold Answer}
&
\textcolor{red}{``Nodality'',``Guidewire Software'',``Jamba Juice'',``Megapath'',``Apple Inc.'',``Yelp'',``Micron Technology'',``Amec Foster Wheeler'',``Genomic Health Inc.'',``Juniper Networks'',``Digital Chocolate'',``Gilt Groupe'',``Hi5 Networks, Inc.'',``Genzyme'',``Sulake'', ...}

\\
\hline

\end{tabular}
\end{adjustbox}

\normalsize
\caption{SFT trajectory example on GrailQA.}
\label{tab:sft-grailqa}
\end{table*}

\begin{table*}
\centering
\small
\setlength{\tabcolsep}{4pt}
\begin{adjustbox}{max width=\textwidth}
\begin{tabular}{
  >{\raggedright\arraybackslash}p{0.13\textwidth}
  >{\raggedright\arraybackslash}p{0.82\textwidth}
}
\toprule

\textbf{Dataset}
&
CWQ
\\
\midrule

\textbf{Question}
&
Lou Seal is the mascot for the team that last won the World Series when?
\\
\midrule

\textbf{Answer}
&
\textbf{Gold:} \textcolor{blue}{2014 World Series};
\textbf{Pred:} \textcolor{blue}{2014 World Series}
\\
\midrule

\textbf{Turn 1}
&
\textbf{Assistant:}

\textcolor{violet}{$<$think$>$}I need to inspect incoming relations for ``Lou Seal''.\textcolor{violet}{$<$/think$>$}

\textcolor{cyan}{$<$search$>$}\{``action'':``get\_head\_relations'',``entity'':``Lou Seal''\}\textcolor{cyan}{$<$/search$>$}

\textbf{Observation:}
\{``relations'':[``\seqsplit{sports.sports\_team.team\_mascot}'']\}
\\
\midrule

\textbf{Turn 2}
&
\textbf{Assistant:}

\textcolor{violet}{$<$think$>$}The relation ``sports.sports\_team.team\_mascot'' is relevant, so I will follow it from ``Lou Seal''.\textcolor{violet}{$<$/think$>$}

\textcolor{cyan}{$<$search$>$}\{``action'':``get\_head\_entities'',``entity'':``Lou Seal'',``relation'':``sports.sports\_team.team\_mascot''\}\textcolor{cyan}{$<$/search$>$}

\textbf{Observation:}
\{``entities'':[``San Francisco Giants'']\}
\\
\midrule

\textbf{Turn 3}
&
\textbf{Assistant:}

\textcolor{violet}{$<$think$>$}I need to inspect outgoing relations for ``San Francisco Giants''.\textcolor{violet}{$<$/think$>$}

\textcolor{cyan}{$<$search$>$}\{``action'':``get\_tail\_relations'',``entity'':``San Francisco Giants''\}\textcolor{cyan}{$<$/search$>$}

\textbf{Observation:}
\{``relations'':[``\seqsplit{sports.sports\_team.team\_mascot}'',
``\seqsplit{baseball.baseball\_team.division}'',
``\seqsplit{sports.sports\_team.championships}'',
``\seqsplit{sports.sports\_team.location}'',
``\seqsplit{sports.sports\_team.founded}'', ...]\}
\\
\midrule

\textbf{Turn 4}
&
\textbf{Assistant:}

\textcolor{violet}{$<$think$>$}The relation ``sports.sports\_team.championships'' is relevant, so I will follow it from ``San Francisco Giants''.\textcolor{violet}{$<$/think$>$}

\textcolor{cyan}{$<$search$>$}\{``action'':``get\_tail\_entities'',``entity'':``San Francisco Giants'',``relation'':``sports.sports\_team.championships''\}\textcolor{cyan}{$<$/search$>$}

\textbf{Observation:}
\{``entities'':[``2010 World Series'', ``2012 World Series'', \textcolor{blue}{``2014 World Series''}]\}
\\
\midrule

\textbf{Final Answer}
&
\textcolor{magenta}{$<$answer$>$}\{``answers'':[``2014 World Series'']\}\textcolor{magenta}{$<$/answer$>$}
\\

\bottomrule
\end{tabular}
\end{adjustbox}
\normalsize
\caption{OACR inference trajectory on CWQ.}
\label{tab:oacr_infer_cwq}
\end{table*}

\begin{table*}
\centering
\small
\setlength{\tabcolsep}{4pt}
\begin{adjustbox}{max width=\textwidth}
\begin{tabular}{
  >{\raggedright\arraybackslash}p{0.13\textwidth}
  >{\raggedright\arraybackslash}p{0.82\textwidth}
}
\toprule

\textbf{Dataset}
&
WebQSP
\\
\midrule

\textbf{Question}
&
what does jamaican people speak?
\\
\midrule

\textbf{Answer}
&
\textbf{Gold:} \textcolor{blue}{Jamaican English; Jamaican Creole English Language};
\textbf{Pred:} \textcolor{blue}{Jamaican Creole English Language; Jamaican English}
\\
\midrule

\textbf{Turn 1}
&
\textbf{Assistant:}

\textcolor{violet}{$<$think$>$}I need to inspect outgoing relations for ``Jamaica''.\textcolor{violet}{$<$/think$>$}

\textcolor{cyan}{$<$search$>$}\{``action'':``get\_tail\_relations'',``entity'':``Jamaica''\}\textcolor{cyan}{$<$/search$>$}

\textbf{Observation:}
\{``relations'':[``\seqsplit{location.country.languages\_spoken}'',
``\seqsplit{location.country.official\_language}'',
``\seqsplit{location.location.people\_born\_here}'',
``\seqsplit{location.country.national\_anthem}'',
``\seqsplit{location.country.capital}'', ...]\}
\\
\midrule

\textbf{Turn 2}
&
\textbf{Assistant:}

\textcolor{violet}{$<$think$>$}The relation ``location.country.languages\_spoken'' is relevant, so I will follow it from ``Jamaica''.\textcolor{violet}{$<$/think$>$}

\textcolor{cyan}{$<$search$>$}\{``action'':``get\_tail\_entities'',``entity'':``Jamaica'',``relation'':``\seqsplit{location.country.languages\_spoken}''\}\textcolor{cyan}{$<$/search$>$}

\textbf{Observation:}
\{``entities'':[\textcolor{blue}{``Jamaican Creole English Language''}, \textcolor{blue}{``Jamaican English''}]\}
\\
\midrule

\textbf{Final Answer}
&
\textcolor{magenta}{$<$answer$>$}\{``answers'':[``Jamaican Creole English Language'',``Jamaican English'']\}\textcolor{magenta}{$<$/answer$>$}
\\

\bottomrule
\end{tabular}
\end{adjustbox}
\normalsize
\caption{OACR inference trajectory on WebQSP.}
\label{tab:oacr_infer_webqsp}
\end{table*}

\begin{table*}
\centering
\small
\setlength{\tabcolsep}{4pt}
\begin{adjustbox}{max width=\textwidth}
\begin{tabular}{
  >{\raggedright\arraybackslash}p{0.13\textwidth}
  >{\raggedright\arraybackslash}p{0.82\textwidth}
}
\toprule

\textbf{Dataset}
&
GrailQA
\\
\midrule

\textbf{Question}
&
which play is produced by the illusion?
\\
\midrule

\textbf{Answer}
&
\textbf{Gold:} \textcolor{blue}{The Illusion};
\textbf{Pred:} \textcolor{blue}{The Illusion}
\\
\midrule

\textbf{Turn 1}
&
\textbf{Assistant:}

\textcolor{violet}{$<$think$>$}I need to inspect outgoing relations for ``The Illusion''.\textcolor{violet}{$<$/think$>$}

\textcolor{cyan}{$<$search$>$}\{``action'':``get\_tail\_relations'',``entity'':``The Illusion''\}\textcolor{cyan}{$<$/search$>$}

\textbf{Observation:}
\{``relations'':[``\seqsplit{theater.theater\_production.play}'',
``\seqsplit{theater.theater\_production.date\_opened}'',
``\seqsplit{theater.theater\_production.producing\_company}'',
``\seqsplit{theater.theater\_production.director}'',
``\seqsplit{theater.theater\_production.theater\_production\_venue\_relationship}'',
``\seqsplit{theater.theater\_production.designers}'',
``\seqsplit{award.award\_nominated\_work.award\_nominations}'']\}
\\
\midrule

\textbf{Turn 2}
&
\textbf{Assistant:}

\textcolor{violet}{$<$think$>$}The relation ``theater.theater\_production.play'' is relevant, so I will follow it from ``The Illusion''.\textcolor{violet}{$<$/think$>$}

\textcolor{cyan}{$<$search$>$}\{``action'':``get\_tail\_entities'',``entity'':``The Illusion'',``relation'':``theater.theater\_production.play''\}\textcolor{cyan}{$<$/search$>$}

\textbf{Observation:}
\{``entities'':[\textcolor{blue}{``The Illusion''}]\}
\\
\midrule

\textbf{Final Answer}
&
\textcolor{magenta}{$<$answer$>$}\{``answers'':[``The Illusion'']\}\textcolor{magenta}{$<$/answer$>$}
\\

\bottomrule
\end{tabular}
\end{adjustbox}
\normalsize
\caption{OACR inference trajectory on GrailQA.}
\label{tab:oacr_infer_grailqa}
\end{table*}
\begin{table*}
\centering
\small
\setlength{\tabcolsep}{4pt}

\begin{adjustbox}{max width=\textwidth}
\begin{tabular}{@{}lcccccccc@{}}
\toprule
\multirow{2}{*}{\textbf{Method}}
&
\multicolumn{2}{c}{\textbf{I.I.D.}}
&
\multicolumn{2}{c}{\textbf{Compositional}}
&
\multicolumn{2}{c}{\textbf{Zero-shot}}
&
\multicolumn{2}{c}{\textbf{Overall}}
\\
\cmidrule(lr){2-3}
\cmidrule(lr){4-5}
\cmidrule(lr){6-7}
\cmidrule(lr){8-9}
&
\textbf{F1}
&
\textbf{Hit@1}
&
\textbf{F1}
&
\textbf{Hit@1}
&
\textbf{F1}
&
\textbf{Hit@1}
&
\textbf{F1}
&
\textbf{Hit@1}
\\
\midrule

\multicolumn{9}{c}{\textit{Training-free Methods}} \\

Qwen2.5-3B
& 13.5 & 15.8
& 6.7  & 8.6
& 12.4 & 14.8
& 11.6 & 13.8
\\

Qwen2.5-7B
& 9.9  & 12.5
& 8.8  & 11.6
& 16.8 & 18.9
& 13.6 & 15.9
\\

Llama-3.1-8B
& 14.4 & 17.5
& 10.0 & 12.6
& 21.9 & 26.0
& 17.7 & 21.3
\\

ToG~\citep{tog}
& -- & 79.4
& -- & 67.3
& -- & 86.5
& -- & 81.4
\\

PoG~\citep{pog}
& -- & 87.9
& -- & 69.7
& -- & 88.6
& -- & 84.7
\\

\midrule
\multicolumn{9}{c}{\textit{Training-based Methods}} \\

\multicolumn{9}{c}{
\textit{\textemdash\ backbone: Qwen2.5-3B \textemdash}
} \\

OACR$_{\mathrm{SFT}}$
& 78.7 & 86.7
& 53.2 & 62.6
& 76.8 & 81.5
& 72.6 & 79.0
\\

\textbf{OACR}
& 82.0 & 89.0
& 57.3 & 68.7
& 79.0 & 84.3
& 75.4 & 82.3
\\

\midrule
\multicolumn{9}{c}{
\textit{\textemdash\ backbone: Qwen2.5-7B \textemdash}
} \\

KnowCoder-A1~\citep{KnowCoder-A1}
& 81.1 & --
& 70.6 & --
& --   & --
& --   & 80.5
\\

OACR$_{\mathrm{SFT}}$
& 79.6 & 85.4
& 53.2 & 63.6
& 81.2 & 85.8
& 75.3 & 81.3
\\

\textbf{OACR}
& 85.8 & 90.2
& 68.8 & 77.0
& 86.0 & 91.1
& 82.5 & 88.1
\\

\midrule
\multicolumn{9}{c}{
\textit{\textemdash\ backbone: Llama-3.1-8B \textemdash}
} \\

KBQA-o1~\citep{kbqa-o1}
& 85.5 & --
& 77.6 & --
& 76.1 & --
& 78.5 & --
\\

OACR$_{\mathrm{SFT}}$
& 81.6 & 87.9
& 53.4 & 68.2
& 81.6 & 85.6
& 76.0 & 82.7
\\

\textbf{OACR}
& 87.5 & 92.1
& 68.2 & 79.8
& 86.7 & 91.1
& 83.2 & 89.1
\\

\bottomrule
\end{tabular}
\end{adjustbox}

\normalsize
\caption{
Detailed performance comparison on the I.I.D., compositional,
zero-shot, and overall splits of GrailQA.
}
\label{tab:grailqa_split_results}
\end{table*}

\begin{table*}
\centering
\small
\setlength{\tabcolsep}{4pt}
\begin{adjustbox}{max width=\textwidth}
\begin{tabular}{
  >{\raggedright\arraybackslash}p{0.12\textwidth}
  >{\raggedright\arraybackslash}p{0.83\textwidth}
}
\hline

\textbf{Dataset}
&
GrailQA 
\\
\hline

\textbf{Question}
&
30 geminorum is a star in which constellation?
\\
\hline

\textbf{Answer}
&
\textbf{Gold:} \textcolor{blue}{Gemini};
\textbf{Pred:} \textcolor{red}{Pollux, Mu Geminorum, 30 Geminorum, Rho Geminorum, Epsilon Geminorum, \ldots}
\\
\hline

\textbf{Reasoning Trajectory}
&
\textbf{Turn 1 -- Assistant}

\textcolor{violet}{$<$think$>$}I need to inspect incoming relations for ``30 Geminorum''.\textcolor{violet}{$<$/think$>$}

\textcolor{cyan}{$<$search$>$}\{``action'':``get\_head\_relations'',``entity'':``30 Geminorum''\}\textcolor{cyan}{$<$/search$>$}

\textbf{Observation}

\{``relations'':[``\seqsplit{astronomy.constellation.contains}'']\}

\medskip

\textbf{Turn 2 -- Assistant}

\textcolor{violet}{$<$think$>$}The relation ``astronomy.constellation.contains'' is relevant, so I will follow it from ``30 Geminorum''.\textcolor{violet}{$<$/think$>$}

\textcolor{cyan}{$<$search$>$}\{``action'':``get\_head\_entities'',``entity'':``30 Geminorum'',``relation'':``astronomy.constellation.contains''\}\textcolor{cyan}{$<$/search$>$}

\textbf{Observation}

\{``entities'':[\textcolor{blue}{``Gemini''}]\}

\medskip

\textbf{Turn 3 -- Assistant}

\textcolor{violet}{$<$think$>$}I need to inspect outgoing relations for ``Gemini''.\textcolor{violet}{$<$/think$>$}

\textcolor{cyan}{$<$search$>$}\{``action'':``get\_tail\_relations'',``entity'':``Gemini''\}\textcolor{cyan}{$<$/search$>$}

\textbf{Observation}

\{``relations'':[``\seqsplit{astronomy.constellation.bordering\_constellations\_new}'',
``\seqsplit{astronomy.constellation.meteor\_showers}'',
``\seqsplit{astronomy.constellation.contains}'',
``\seqsplit{astronomy.constellation.galaxies\_observed\_in\_constellation}'']\}

\medskip

\textbf{Turn 4 -- Assistant}

\textcolor{violet}{$<$think$>$}The relation ``astronomy.constellation.contains'' is relevant, so I will follow it from ``Gemini''.\textcolor{violet}{$<$/think$>$}

\textcolor{cyan}{$<$search$>$}\{``action'':``get\_tail\_entities'',``entity'':``Gemini'',``relation'':``\seqsplit{astronomy.constellation.contains}''\}\textcolor{cyan}{$<$/search$>$}

\textbf{Observation}

\{``entities'':[\textcolor{red}{``Pollux''}, \textcolor{red}{``Mu Geminorum''}, ``30 Geminorum'',
\textcolor{red}{``Rho Geminorum''}, ``Gemini'', \textcolor{red}{``Epsilon Geminorum''},
``Gamma Geminorum'', ``Delta Geminorum'', ``Zeta Geminorum'', ``Tau Geminorum'', \ldots]\}
\\
\midrule

\textbf{Failure Point}
&
The gold answer \textcolor{blue}{Gemini} is already obtained in Turn 2. However, the model does not stop and instead continues to expand from \textcolor{blue}{Gemini} through \texttt{astronomy.constellation.contains}. This extra hop changes the target from the constellation to the stars contained in that constellation, leading to an over-expanded answer list rather than the required constellation name.
\\
\hline

\end{tabular}
\end{adjustbox}
\normalsize
\caption{Case study of \textsc{Answer Mismatch}.}
\label{tab:case_answer_mismatch}
\end{table*}

\begin{table*}
\centering
\small
\setlength{\tabcolsep}{4pt}
\begin{adjustbox}{max width=\textwidth}
\begin{tabular}{
  >{\raggedright\arraybackslash}p{0.12\textwidth}
  >{\raggedright\arraybackslash}p{0.83\textwidth}
}
\hline

\textbf{Dataset}
&
CWQ 
\\
\hline
\textbf{Question}
&
What organization founded after 1552 is the school JRR Tolkien attended?
\\
\hline

\textbf{Answer}
&
\textbf{Gold:} \textcolor{blue}{St. Philip's School};
\textbf{Pred:} \textcolor{red}{University of Oxford}
\\
\hline

\textbf{Reasoning Trajectory}
&
\textbf{Turn 1 -- Assistant}

\textcolor{violet}{$<$think$>$}I need to inspect outgoing relations for ``J. R. R. Tolkien''.\textcolor{violet}{$<$/think$>$}

\textcolor{cyan}{$<$search$>$}\{``action'':``get\_tail\_relations'',``entity'':``J. R. R. Tolkien''\}\textcolor{cyan}{$<$/search$>$}

\textbf{Observation}

\{``relations'':[``\texttt{people.person.education}'', ``organization.organization\_member.member\_of'', ``organization.organization\_founder.organizations\_founded'', ``book.author.openlibrary\_id'', ...]\}

\medskip

\textbf{Turn 2 -- Assistant}

\textcolor{violet}{$<$think$>$}The relation ``people.person.education'' is relevant, so I will follow it from ``J. R. R. Tolkien''.\textcolor{violet}{$<$/think$>$}

\textcolor{cyan}{$<$search$>$}\{``action'':``get\_tail\_entities'',``entity'':``J. R. R. Tolkien'',``relation'':``people.person.education''\}\textcolor{cyan}{$<$/search$>$}

\textbf{Observation}

\{``entities'':[``King Edward's School, Birmingham'', ``J. R. R. Tolkien'', \textcolor{blue}{``St. Philip's School''}, \textcolor{red}{``University of Oxford''}, ``Exeter College, Oxford'', ``First Class Honours'', ``1915'', ``English Language'', ``1911-10'']\}
\\
\midrule

\textbf{Failure Point}
&
The gold answer \textcolor{blue}{St. Philip's School} is already retrieved in Turn 2, but the model finally selects \textcolor{red}{University of Oxford}. The failure comes from not applying the question constraint \emph{founded after 1552} when choosing among the retrieved education candidates.
\\
\hline

\end{tabular}
\end{adjustbox}
\normalsize
\caption{Case study of \textsc{Constraint Failure}.}
\label{tab:case_constraint_failure}
\end{table*}

\begin{table*}
\centering
\small
\setlength{\tabcolsep}{4pt}
\begin{adjustbox}{max width=\textwidth}
\begin{tabular}{
  >{\raggedright\arraybackslash}p{0.12\textwidth}
  >{\raggedright\arraybackslash}p{0.83\textwidth}
}
\hline

\textbf{Dataset}
&
WebQSP 
\\
\hline

\textbf{Question}
&
which is the main train station in rome?
\\
\hline

\textbf{Answer}
&
\textbf{Gold:} \textcolor{blue}{Roma Termini railway station};
\textbf{Pred:} \textcolor{red}{Rome Metro}
\\
\hline

\textbf{Reasoning Trajectory}
&
\textbf{Turn 1 -- Assistant}

\textcolor{violet}{$<$think$>$}I need to inspect outgoing relations for ``Rome''.\textcolor{violet}{$<$/think$>$}

\textcolor{cyan}{$<$search$>$}\{``action'':``get\_tail\_relations'',``entity'':``Rome''\}\textcolor{cyan}{$<$/search$>$}

\textbf{Observation}

\{``relations'':[``\seqsplit{travel.travel\_destination.how\_to\_get\_here}'',
\textcolor{red}{``\seqsplit{travel.travel\_destination.local\_transportation}''},
``\seqsplit{travel.travel\_destination.tourist\_attractions}'',
``\seqsplit{location.location.contains}'',
``\seqsplit{location.location.nearby\_airports}'',
``\seqsplit{location.dated\_location.date\_founded}'', ...]\}

\medskip

\textbf{Turn 2 -- Assistant}

\textcolor{violet}{$<$think$>$}The relation ``travel.travel\_destination.local\_transportation'' is relevant, so I will follow it from ``Rome''.\textcolor{violet}{$<$/think$>$}

\textcolor{cyan}{$<$search$>$}\{``action'':``get\_tail\_entities'',``entity'':``Rome'',``relation'':``\seqsplit{travel.travel\_destination.local\_transportation}''\}\textcolor{cyan}{$<$/search$>$}

\textbf{Observation}

\{``entities'':[``ATAC SpA'', \textcolor{red}{``Rome Metro''}]\}
\\
\midrule

\textbf{Failure Point}
&
The model selects the broad relation \textcolor{red}{\texttt{travel.travel\_destination.local\_transportation}}, which leads to local transportation systems such as \textcolor{red}{Rome Metro}. However, the question asks for the main train station in Rome, whose gold answer is \textcolor{blue}{Roma Termini railway station}. The error therefore comes from choosing an incorrect relation/path before answer selection, rather than from tool execution failure.
\\
\hline

\end{tabular}
\end{adjustbox}
\normalsize
\caption{Case study of \textsc{Path Error}.}
\label{tab:case_path_error}
\end{table*}

\end{document}